\PassOptionsToPackage{shortlabels}{enumitem}
\PassOptionsToPackage{table,xcdraw}{xcolor}

\documentclass[11pt, a4paper]{include/gdm_format}

\usepackage[authoryear, sort&compress, round]{natbib}
\usepackage{subcaption}
\usepackage{mwe}
\usepackage{graphicx}
\usepackage{xr}
\usepackage{marvosym}
\usepackage{makecell}

\usepackage{enumitem} %

\usepackage{fvextra}

\usepackage{amsmath,amsfonts,bm}

\def\eqref#1{equation~\ref{#1}}

\def\1{\bm{1}}

\DeclareMathAlphabet{\mathsfit}{\encodingdefault}{\sfdefault}{m}{sl}
\SetMathAlphabet{\mathsfit}{bold}{\encodingdefault}{\sfdefault}{bx}{n}

\usepackage{tablefootnote}
\usepackage{pdfpages}
\usepackage{minitoc}
\usepackage{titletoc}
\usepackage{appendix}

\usepackage{hyperref}
\usepackage{url}
\usepackage{xurl}
\usepackage[parfill]{parskip}

\usepackage{amsmath}
\usepackage{xspace}
\usepackage{multirow}
\usepackage{booktabs}
\usepackage{color}
\usepackage{colortbl}
\usepackage{cleveref}
\usepackage{graphicx}
\usepackage{algorithm}
\usepackage{algorithmicx}
\usepackage{algpseudocode}
\usepackage{soul}
\usepackage{multicol}
\usepackage{pifont}
\usepackage[normalem]{ulem}
\usepackage{listings}
\usepackage{adjustbox}
\usepackage{pifont}

\newcommand{\cmark}{\ding{51}} %
\newcommand{\xmark}{\ding{55}} %

\title{\ouralgo: Workshop-Level Automated Scientific Discovery via Agentic Tree Search}

\correspondingauthor{Yutaro Yamada (yutaroyamada@sakana.ai), Robert Tjarko Lange (robert@sakana.ai), and Cong Lu (cong@sakana.ai).}

\author[1,*]{Yutaro Yamada}
\author[1,*]{Robert Tjarko Lange}
\author[1,2,3,*]{Cong Lu}
\author[1,2,3]{Shengran Hu}
\author[4]{Chris Lu}
\author[4]{Jakob Foerster}
\author[2,3,5,\Cross]{Jeff Clune}
\author[1,\Cross]{David Ha}

\affil[*]{Equal Contribution}
\affil[1]{Sakana AI}
\affil[2]{University of British Columbia}
\affil[3]{Vector Institute}
\affil[4]{FLAIR, University of Oxford}
\affil[5]{Canada CIFAR AI Chair}
\affil[\Cross]{Equal Advising}

\newcommand{\oldalgo}{\textsc{The AI Scientist-v1}\xspace}
\newcommand{\ouralgo}{\textsc{The AI Scientist-v2}\xspace}

\usepackage{tcolorbox}
\tcbuselibrary{breakable}
\tcbuselibrary{listings}
\definecolor{lightgreen}{RGB}{235, 255, 235}

\definecolor{codegreen}{rgb}{0,0.6,0}
\definecolor{codegray}{rgb}{0.5,0.5,0.5}
\definecolor{codepurple}{rgb}{0.58,0,0.82}
\definecolor{backcolour}{RGB}{245,248,250}
\definecolor{emph}{RGB}{166,88,53}
\definecolor{nightblue}{RGB}{9,49,105}
\definecolor{keywords}{RGB}{207,33,46}
\definecolor{lightpurple}{RGB}{130,81,223}

\lstdefinestyle{mystyle}{
    backgroundcolor=\color{backcolour},   
    commentstyle=\color{codegreen},
    keywordstyle=\color{keywords},
    stringstyle=\color{nightblue},
    basicstyle=\ttfamily\scriptsize,
    breakatwhitespace=false,         
    breaklines=true,                 
    captionpos=b,                    
    keepspaces=true,                 
    showspaces=false,                
    showstringspaces=false,
    showtabs=false,                  
    tabsize=2,
    frame=shadowbox,
    emph={AutoTokenizer,AutoModelForSequenceClassification,Explainer},
    emphstyle={\color{emph}},
    emph={[2]from_pretrained,compute_table},
    emphstyle={[2]\color{lightpurple}},
}

\begin{document}

\begin{abstract}
AI is increasingly playing a pivotal role in transforming how scientific discoveries are made.
We introduce \ouralgo, an end-to-end agentic system capable of producing the first entirely AI-generated peer-review-accepted workshop paper.
This system iteratively formulates scientific hypotheses, designs and executes experiments, analyzes and visualizes data, and autonomously authors scientific manuscripts.
Compared to its predecessor~\citep[v1,][]{lu2024ai}, \ouralgo eliminates the reliance on human-authored code templates, generalizes effectively across diverse machine learning domains, and leverages a novel progressive agentic tree-search methodology managed by a dedicated experiment manager agent.
Additionally, we enhance the AI reviewer component by integrating a Vision-Language Model (VLM) feedback loop for iterative refinement of content and aesthetics of the figures.
We evaluated \ouralgo by submitting three fully autonomous manuscripts to a peer-reviewed ICLR workshop.
Notably, one manuscript achieved high enough scores to exceed the average human acceptance threshold, marking the first instance of a fully AI-generated paper successfully navigating a peer review.
This accomplishment highlights the growing capability of AI in conducting all aspects of scientific research.
We anticipate that further advancements in autonomous scientific discovery technologies will profoundly impact human knowledge generation, enabling unprecedented scalability in research productivity and significantly accelerating scientific breakthroughs, greatly benefiting society at large.
We have open-sourced the code at \url{https://github.com/SakanaAI/AI-Scientist-v2} to foster the future development of this transformative technology. We also discuss the role of AI in science, including AI safety.
\end{abstract}

\maketitle
\section{Introduction}
 
Automated scientific discovery empowered by artificial intelligence (AI) has garnered considerable attention in recent years~\citep{Wang2023scientificdiscovery, king2009doi:10.1126/science.1165620, gil2014doi:10.1126/science.1259439, Kitano2021, Xu2021, Cornelio2023}.
The development of end-to-end frameworks capable of autonomously formulating hypotheses, performing experiments, analyzing results, and authoring manuscripts could fundamentally transform the scientific process.
A notable recent advance in this direction is \oldalgo~\citep{lu2024ai}, which demonstrated the feasibility of a fully automated scientific workflow and downstream manuscript production.
However, significant limitations constrained its broad applicability and autonomy.
Specifically, it relied heavily on human-authored code templates requiring manual effort to create a new template for each new topic area.
Furthermore, its linear and shallow experimentation approach prevented deeper exploration of scientific hypotheses.

\begin{table}[htbp]
    \centering
    \caption{\textbf{Comparison of AI Scientist Versions.} Comparison highlights key advancements in \ouralgo, including autonomous code generation via tree search, enhanced VLM integration for feedback during experiments and manuscript review, and evaluation through formal peer review.}
    \label{tab:ai-scientist-comparison}
    \begin{adjustbox}{width=\textwidth}
    \begin{tabular}{lcccccc}
        \toprule
\textit{Feature} & \textbf{Codebase} & \textbf{Execution} & \textbf{Parallel} & \textbf{VLM} & \textbf{Human Result} \\
        & \textbf{Drafting} & \textbf{Planning} & \textbf{Experiments} & \textbf{Reviewer}  & \textbf{Evaluation} \\
        \midrule
        \oldalgo & Topic-Specific & Linear & \xmark & \xmark & Not Submitted\\
        \rowcolor{gray!20}
        \ouralgo & Domain-General & Tree-Based & \cmark & \cmark & Workshop Acceptance-Worthy\\
        \bottomrule
    \end{tabular}
    \end{adjustbox}
\end{table}

In this paper, we introduce \ouralgo, a substantially improved successor that directly addresses these limitations.
Our contributions are threefold.
First, we eliminate the dependency on human-provided code templates, significantly increasing the system's autonomy and ability to be deployed out of the box across multiple machine learning domains.
Second, we introduce an experiment manager agent coupled with a novel agentic tree-search algorithm, enabling deeper and more systematic exploration of complex hypotheses.
Third, we enhance the reviewing and refinement stages by integrating a Vision-Language Model (VLM)-based feedback mechanism, improving the quality, clarity, and alignment of generated figures, captions, and text interpretation.
To rigorously evaluate the capabilities and limitations of fully autonomous manuscript generation, we conducted a controlled experiment: three manuscripts entirely generated by \ouralgo were submitted to a peer-reviewed workshop at ICLR.
Remarkably, one manuscript achieved an average reviewer score of 6.33 (placing it roughly in the top 45\% of submissions) and would have been accepted after meta-review were it human-generated, thus becoming the first fully AI-generated manuscript to successfully pass a peer-review process. 

The accepted paper investigates whether incorporating an explicit compositional regularization term into neural network training can improve compositional generalization. Specifically, it penalizes large deviations between embeddings of successive time steps in sequence models, hypothesizing that this encourages compositionality. The approach is evaluated using synthetic arithmetic expression datasets, but it is found that compositional regularization does not yield significant improvements and occasionally harms performance.
The workshop reviewers appreciated the paper for clearly identifying the challenges of effective compositional regularization and reporting on negative results. However, they collectively highlighted shortcomings, including insufficient justification and intuitive explanations for why the chosen regularization method would enhance compositionality.
Our personal assessment (detailed further in \S\ref{sec:humaneval}) highlights several additional potential improvements in method description (e.g., making clear exactly which component of the network is being regularized), potential dataset overlap issues, and inaccuracies in figure captions. Overall, reviewers viewed the paper as an interesting and technically sound workshop contribution that needs further development and broader experimentation to reach conference-level rigor.

This report provides an in-depth outline of the developed methodological advances, analysis of the workshop-submitted papers, and a discussion on the ethical and safety considerations of systems like \ouralgo. Our overall contributions are as follows:

\begin{enumerate}
    \item We introduce \ouralgo, an automated scientific discovery framework enhanced by agentic tree search, VLM feedback, and parallel experiment execution. It thereby significantly improves the autonomy, flexibility, and scientific exploration depth of previous systems.
    \item We demonstrate, for the first time, that an AI-generated manuscript can successfully pass peer review at a recognized machine learning workshop, marking a critical milestone for AI science.
    \item We conduct comprehensive internal evaluations and analyses of both peer-review feedback and our system's outputs, providing insights into the strengths, weaknesses, and current status of AI-generated manuscripts relative to traditional human-authored scientific publications.
    \item We open-source the \href{https://github.com/SakanaAI/AI-Scientist-v2}{full codebase} for \ouralgo and the \href{https://github.com/SakanaAI/AI-Scientist-ICLR2025-Workshop-Experiment/}{ICLR 2025 workshop experiment data}, encouraging further exploration by the research community and advancing a discussion regarding AI's evolving role in science--in the open.
\end{enumerate}

\section{Background}

\begin{figure}[t!]
\centering
\includegraphics[width=0.925\textwidth]{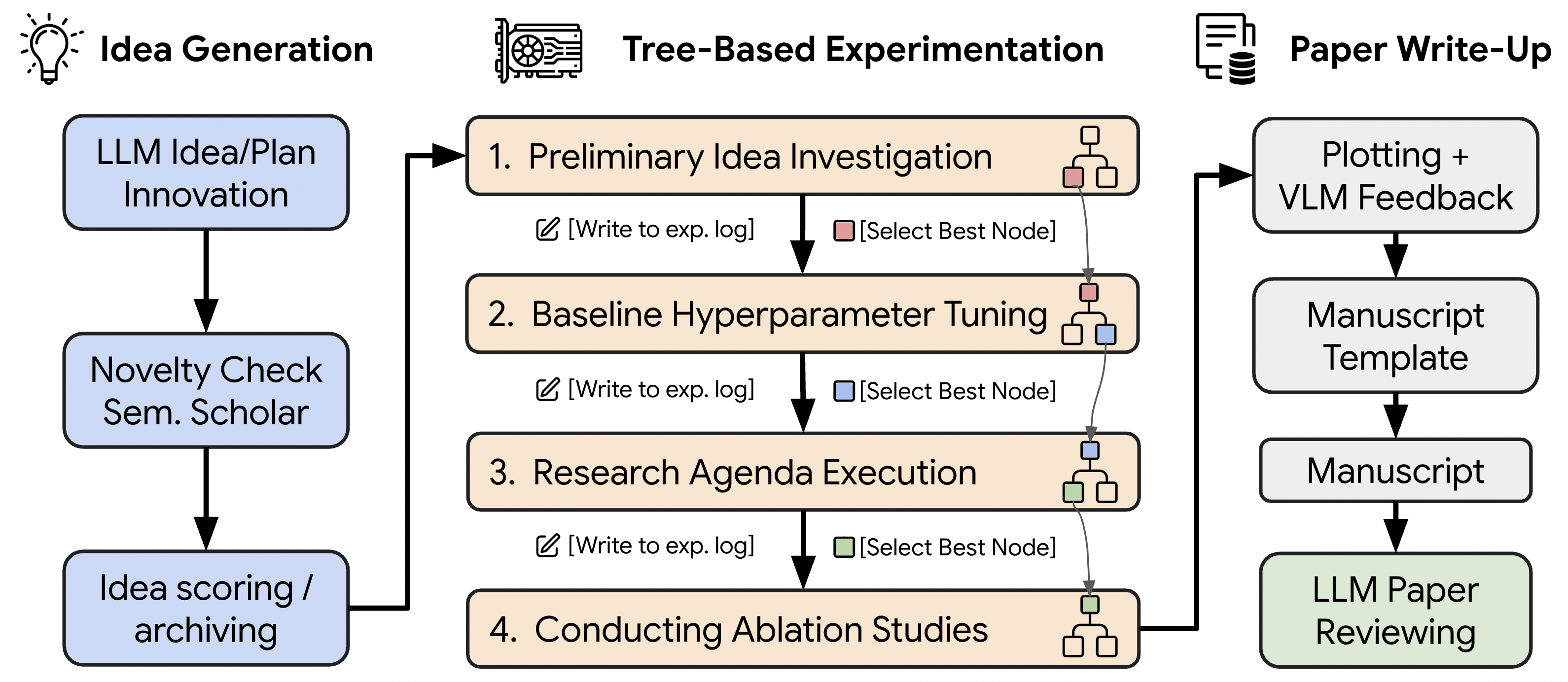}
\caption{\textbf{\ouralgo Workflow.} The workflow consists of several phases covering automated idea generation, experiment execution, figure visualization, manuscript writing, and reviewing. Unlike the initial version, \ouralgo removes the dependency on human-coded templates. Instead, it employs agentic tree search (managed by an Experiment Progress Manager across several stages, orange) to generate and refine code implementations. Subsequent experimentation leverages the best-performing code checkpoints (nodes) from the tree search to iteratively test various research hypotheses.}
\label{fig:conceptual}
\end{figure}

\oldalgo~\citep{lu2024ai} introduced the first AI system that entirely automates scientific discovery and the presentation of its results. Given a baseline code template, it autonomously wrote code, executed experiments, visualized outcomes, and produced a complete scientific manuscript. However, despite representing a significant step forward, \oldalgo was subject to limitations. Foremost among these was its reliance on human-crafted baseline code templates, significantly constraining its autonomy and hindering unconstrained out-of-the-box deployability. Instead, human effort was still required to draft an initial base experiment outline in code. Additionally, the experimentation process followed a strictly linear hypothesis-testing routine, limiting depth and exploration flexibility, especially when addressing complex research questions.

\textbf{Language Model Agent Scaffolding.}
To further enhance LLM performance on complex reasoning tasks, researchers have developed agentic scaffolding frameworks, each with distinct advantages and limitations. For example, Reflexion~\citep{shinn2024reflexion} enables models to iteratively reflect on previous responses, encouraging self-improvement through critical evaluation of past outputs; it improves robustness, but can introduce computational overhead and slower inference. Another promising direction is the integration of tree-search strategies with LLMs~\citep{aide2025}, allowing structured exploration of reasoning paths. This approach enhances systematic reasoning and comprehensiveness, though at the cost of increased complexity, higher computational demands, and challenges in scalability.

\textbf{Tree Search with Large Language Models.}
We empirically observed that automated research conducted by \oldalgo often resulted in short-sighted experimentation. The human-driven scientific process, on the other hand, relies on open-ended hypothesis generation, stepping-stone collection, and iterative hypothesis refinement.
Recent advances using code generation as an action space have opened new opportunities for LLM-driven automated workflows~\citep{wang2024CodeAct}. AIDE~\citep{aide2025} combines LLM-based code generation with tree search, demonstrating state-of-the-art performance on the MLEBench benchmark~\citep{chan2025mlebench}, designed for machine learning engineering tasks. In AIDE, each node represents a potential solution state with a corresponding scalar evaluation score (e.g., validation accuracy). Nodes are iteratively selected for further debugging or refinement based on these scores. Inspired by this approach, we integrate a similar tree search-based exploration strategy within our automated scientific discovery framework, adapting it specifically to the multi-stage nature of scientific experimentation, as detailed in \S\ref{sec:main_aisci_design}.

\section{\ouralgo}
\label{sec:main_aisci_design}

We now describe the major innovations introduced in \ouralgo relative to \oldalgo~\citep{lu2024ai}.
The most significant improvement is the move towards greater autonomy and generalization, starting a more general idea generation phase (\S\ref{sec:generalized_idea_gen}) and eliminating the reliance on fixed, human-authored template code for experimentation.
This process begins with generalized idea generation, producing an initial concept, which then feeds into the experimentation phase (\S\ref{sec:removing_template_dependency}).
To manage this, we introduce two critical features in the experimentation phase: \emph{coarse-grained experiment management} and \emph{agentic tree search-based exploration}.
Additionally, we integrate Vision Language Models (VLMs) into the experimental and review phases (\S\ref{sec:vlm_reviewer}).
Finally, we streamline the manuscript writing phase by replacing the incremental, Aider-based~\citep{gauthier2024aider} iterative writing approach of \oldalgo with a simpler, single-pass generation followed by a separate reflection stage powered by reasoning models such as o1~\citep{OpenAIOS}.
We include a full list of sampling hyperparameters and models used in \Cref{appsec:hypers} and the prompts used for \ouralgo in \Cref{appsec:prompts}.

\subsection{More General Idea Generation}
\label{sec:generalized_idea_gen}
A key conceptual shift in \ouralgo is the approach to research idea generation. Unlike the predecessor system, which primarily focused on proposing incremental modifications or extensions based on an existing codebase, \ouralgo adopts a process that begins at a higher level of abstraction. The system is prompted to engage in more open-ended thinking about potential research directions, hypotheses, and experimental designs, akin to formulating a research abstract or grant proposal before committing to a specific implementation.

This approach encourages the exploration of potentially more novel or foundational ideas, rather than being constrained by the structure and topics of pre-existing code. It aligns more closely with how researchers often develop broader research visions, starting with abstract concepts and assessing novelty and feasibility before diving into specific implementations.
Crucially, this generalized idea generation phase integrates literature review tools, such as Semantic Scholar, in the loop. The system can query the literature database during the idea formulation process to assess the novelty of a proposed concept and identify relevant prior work. This allows for more informed decisions about pursuing a particular research avenue, ensuring ideas are grounded in the existing scientific landscape from the outset, rather than relying solely on post-hoc checks.

\subsection{Removing Template Dependency}
\label{sec:removing_template_dependency}
Following the improved idea generation phase, \ouralgo proceeds with experimentation. Beyond the code-conditioned idea generation, \oldalgo also depended on the predefined template code as a starting baseline implementation. The LLM-driven code changes were then limited to sequential code adaptations. We now outline our strategy for eliminating this limitation, thus improving the system's flexibility and autonomy.

\subsubsection{Experiment Progress Manager}
\label{exp_progress_manager}

Real-world scientific experimentation typically proceeds through distinct stages, from initial feasibility assessments to detailed ablation analyses. To emulate this structured approach, we introduce an \textbf{experiment progress manager agent} that coordinates four clearly defined stages of scientific experimentation:

\begin{enumerate}[label=\textbf{Stage \arabic*}, leftmargin=*]
\item \textbf{Preliminary Investigation}: Establishing initial feasibility and correctness through a minimal working prototype based on the generated research idea.
\item \textbf{Hyperparameter Tuning}: Refining the initial implementation by optimizing critical hyperparameters (e.g., learning rate, epochs) to create a robust experimental baseline.
\item \textbf{Research Agenda Execution}: Systematically implementing the core research agenda based on the tuned baseline.
\item \textbf{Ablation Studies}: Systematically assessing the importance of various research components, providing rigorous support for the main experimental findings.
\end{enumerate}

Each stage has explicit stopping criteria. Stage 1 concludes when a basic working prototype is successfully executed. Stage 2 ends when experiments stabilize, as indicated by convergence in training curves and successful execution across at least two datasets. Stages 3 and 4 conclude when the allocated computational budget is exhausted.
Stage 3 also includes a check for experiment duration--if runs finish much faster than the pre-allocated runtime, the system suggests increasing the complexity of experiments.

After each stage, the experiment manager selects the best-performing node using a dedicated LLM evaluator (see next section) based on clearly articulated criteria.
This selected node is then carried forward to seed the subsequent experimentation stage. The manager also records checkpoints at each stage's completion.
To ensure scientific rigor and reproducibility, the experiment manager launches multiple replications of the selected best experiments at the conclusion of each stage. These repeated runs provide statistics (mean and standard deviation) for figures and reported results.

\begin{figure}
\centering
\includegraphics[width=0.95\textwidth]{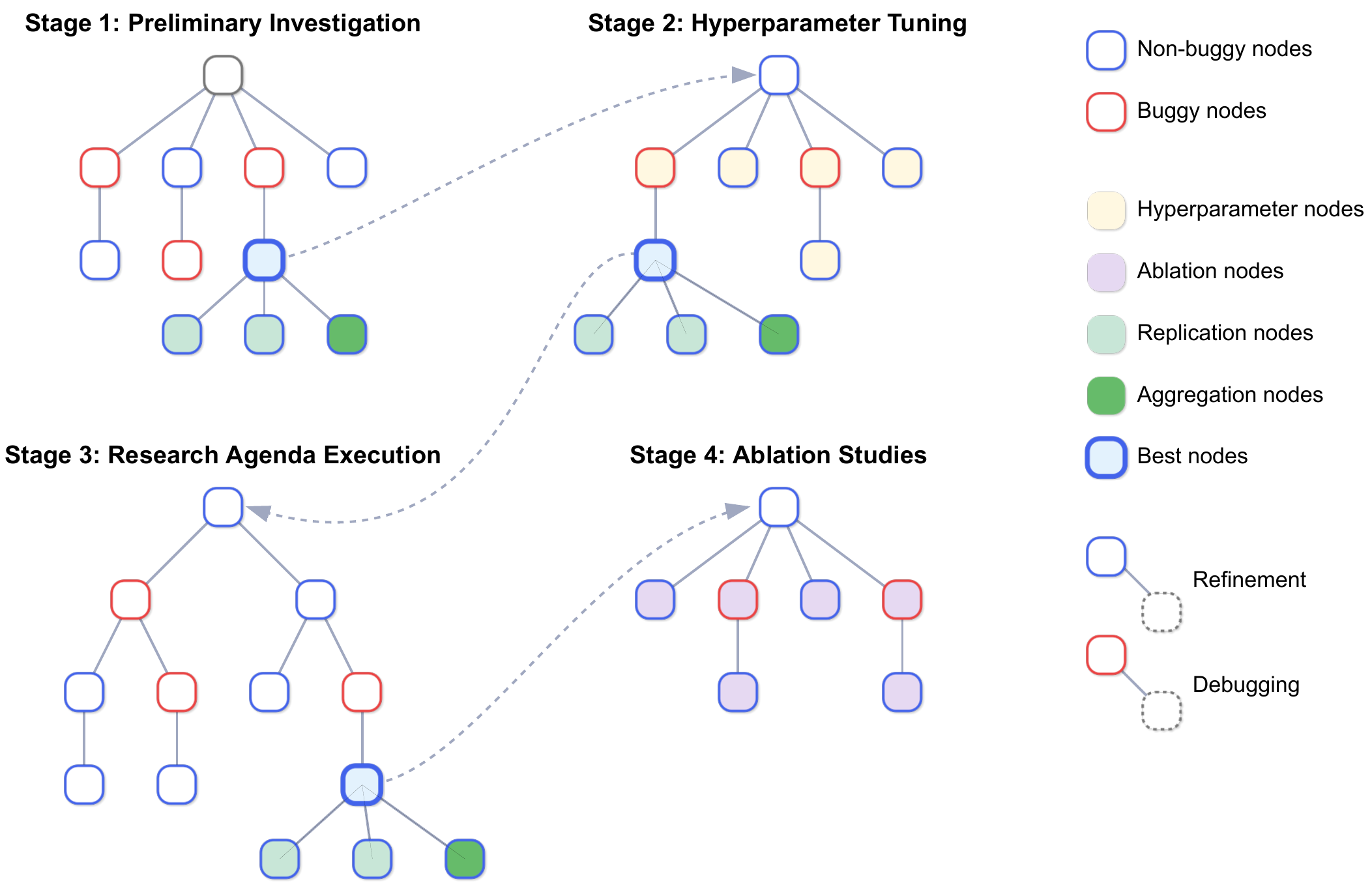}
\caption{\ouralgo workflow showing different stages of tree-based experimentation. 
Stage 1 begins at the root node, where initial experiment code is generated in parallel.
After running the experiment code and visualization scripts, each node is classified based on the outcome: if an error occurs, it is marked as a buggy node; otherwise, it is labeled as a non-buggy node.
New child nodes are created differently depending on their parent node's status:
For non-buggy nodes, refinement is applied to improve the experiment code for better performance.
For buggy nodes, the system attempts to debug them using stored error information.
A best-performing node, selected by LLM-based evaluation, is passed down as the root node of Stage 2.
From this root node, child nodes are created for hyperparameter tuning. 
The top-performing node from Stage 2 is then used to initialize Stage 3, where the system executes the research agenda, applies refinements, and performs debugging as needed.
In Stage 4, similar to Stage 2, the root node generates ablation nodes.
Additionally, replication nodes repeat the same experiment as their parent node, while aggregation nodes collect results from replication nodes to generate combined visualizations and summaries.}
\label{fig:tree_search}
\end{figure}

\subsubsection{Parallelized Agentic Tree Search}
\label{tree-search}

\oldalgo operated strictly linearly, where each code refinement directly built on the immediately preceding experiment. In contrast, \ouralgo adopts a significantly more flexible and exploratory approach inspired by recent successes in integrating tree search with LLM-driven workflows~\citep{aide2025, Wijk2024REBenchEF, chan2025mlebench} and research on open-endedness~\citep{aigas, Mouret2015IlluminatingSS}. We incorporate this agentic tree search approach across all four experimentation stages outlined in \S\ref{exp_progress_manager}, enabling deeper and more systematic exploration of scientific hypotheses.

Each experimental node within our tree-based framework undergoes the following execution cycle: An LLM first generates both a concrete experimentation plan and the associated Python code to implement the experiment. The generated code is immediately executed in a Python interpreter. If execution encounters an error, the error message is recorded, and the node is marked as \textbf{buggy}, ending the current execution cycle for that node. If execution succeeds, the experiment proceeds to the \emph{plotting phase}.

During each experiment, the system is instructed to save all relevant experimental outputs (training and validation metrics, losses, etc.) into structured numpy files. In the plotting phase, \ouralgo reads these stored results and the code, generating visualizations that summarize and illustrate the findings clearly. These visualizations are subsequently passed to a Vision-Language Model (VLM) for critique. Any issues flagged by the VLM (such as unclear labels, missing legends, or misleading visualizations) result in the node being marked as \textbf{buggy}, and this feedback is recorded for future debugging. Nodes that successfully execute and pass the VLM review without issue are designated as \textbf{non-buggy}.

We define each node as a collection comprising an experiment script (e.g., a Python file), a textual description of the high-level plan implemented in the script, an execution error trace (if applicable), experiment runtime, performance metrics recorded during the experiment, feedback from an LLM after running the script, a visualization script, file paths to the generated figures, feedback from a VLM on those figures, and the node's final status (either buggy or non-buggy).

At each iteration, the system selects several nodes from the existing tree to expand in parallel.
With a predefined probability, a \textbf{buggy node} is chosen (thus prioritizing error resolution and debugging); otherwise, a \textbf{non-buggy} node is selected for further refinement and improvement.
When choosing between non-buggy nodes, the system uses a \textbf{best-first search strategy}, guided by an LLM that evaluates candidates based on factors like performance metrics, training dynamics, and the quality of generated plots.
The selected nodes are expanded by creating a new child node that may either attempt debugging if the parent node was buggy, or refine and improve upon the previous experiment if the parent was non-buggy.
An LLM is used to generate the plan and experiment code for each new child node, after which all new nodes are executed concurrently in parallel, significantly accelerating the exploration process.
In addition to buggy and non-buggy nodes, we introduce specialized node variants tailored to specific experimental needs:

\begin{itemize}
\item \textbf{Hyperparameter nodes} systematically explore alternative hyperparameter configurations during Stage 2. The system maintains careful records of previously tested hyperparameters, preventing redundant experiments. Errors encountered during hyperparameter tuning trigger the creation of corresponding debug nodes.
\item \textbf{Ablation nodes} evaluate crucial ablation studies during Stage 4, assessing the importance of various components or assumptions underlying the experiment. Similar to hyperparameter nodes, previously tested ablation conditions are tracked to avoid repetition, and debugging nodes are created in response to any encountered errors.
\item \textbf{Replication nodes} execute replicates of their parent experiments using different random seeds. Typically, several replication nodes are created to enable the calculation of statistical measures (mean and standard deviation) of experimental outcomes, enhancing result robustness.
\item \textbf{Aggregation nodes} are special nodes created to consolidate and visualize the combined results of replication nodes. Unlike other node types, aggregation nodes do not conduct new experiments but simply generate a Python script to aggregate and summarize prior results, producing figures that explicitly show means and standard deviations.
\end{itemize}

The structured design of experiment stages and tailored node types facilitates systematic exploration across all stages.
Unlike some LLM agents that rigidly follow predefined, fine-grained workflow graphs, our approach adopts a looser structure that guides the entire empirical research cycle, enabling flexible system behavior while maintaining coherence across iterative stages.

\subsection{Dataset Loading via Hugging Face}

Most empirical machine learning research relies heavily on publicly available datasets. Hugging Face Hub provides a convenient and unified framework for accessing a wide variety of commonly used datasets, complete with predefined train, validation, and test splits. In \ouralgo, we prompt the system to leverage Hugging Face Hub whenever possible, automatically downloading required datasets using the standard one-line function ({\tt datasets.load\_dataset}).
While this standardized approach greatly simplifies dataset handling, we acknowledge it is somewhat ad-hoc, as not all dataset repositories support this method.
    
\subsection{Vision-Language Model Reviewer}
\label{sec:vlm_reviewer}

Unlike \oldalgo, which did not leverage Vision Language Models (VLMs), \ouralgo incorporates VLMs at two phases of the research workflow:
First, during the tree-based experimentation phase, VLMs provide immediate feedback on generated figures, ensuring that these visualizations effectively and accurately communicate experimental results. Second, during the manuscript writing reflection stage, VLMs evaluate figures and their captions, enhancing the visual clarity and coherence of the resulting paper.

In the paper-writing process, we extract screenshots of figures alongside their captions and the corresponding text from the paper that references them (identified by the keyword ``Figure X''). These images and textual references are then provided to the VLM, which performs multiple quality checks, including verifying the alignment between figures and captions, identifying issues with visual clarity (e.g., missing legends, unclear labels), and detecting potential duplication of figures in the main text and appendix.
Through the iterative integration of VLM feedback, we significantly enhance the visual quality and clarity of manuscripts generated by \ouralgo.

\section{Human Evaluation of Manuscripts Generated by \ouralgo}
\label{sec:humaneval}

\begin{figure}[h!]
\centering
\includegraphics[width=\textwidth]{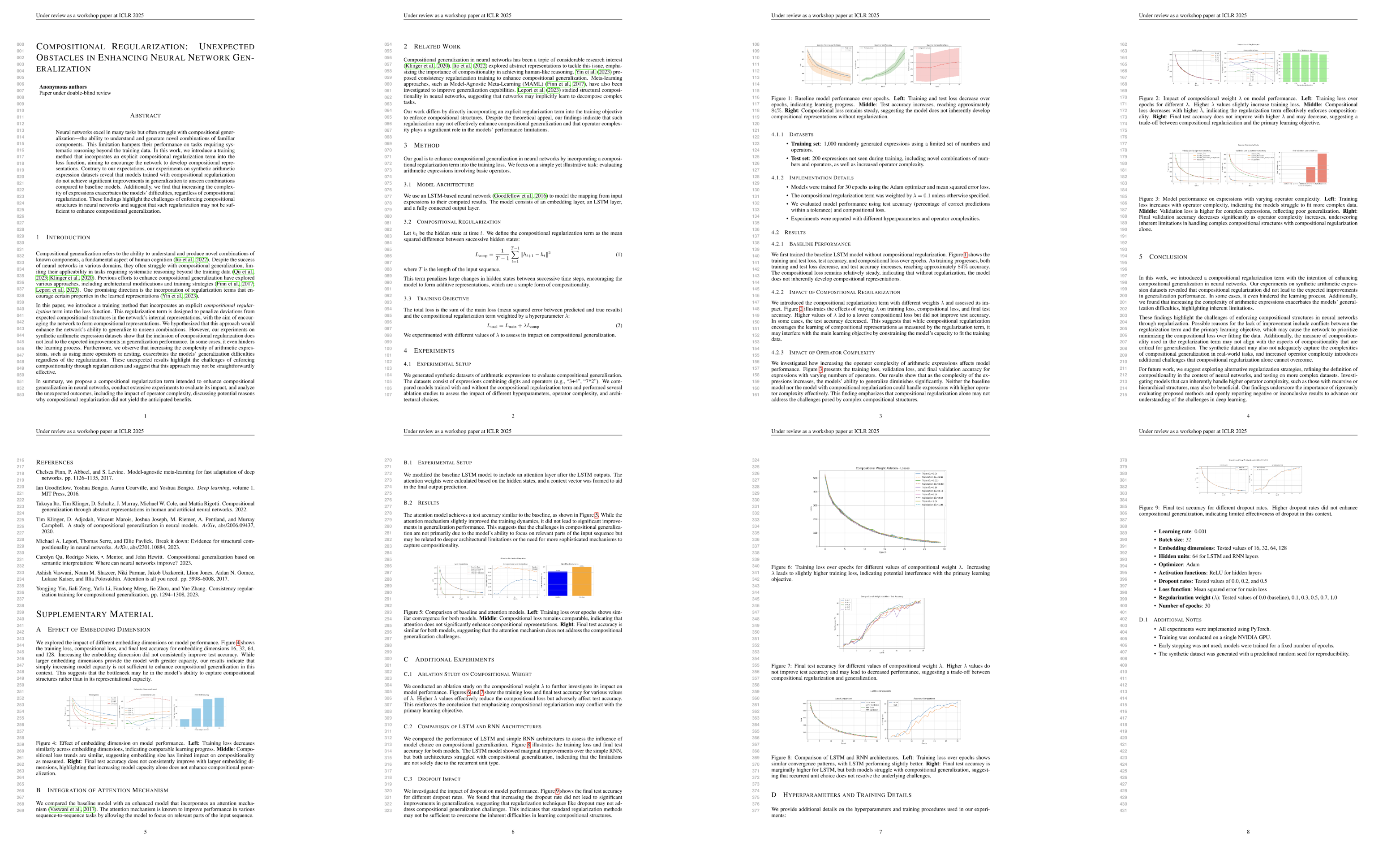}
\caption{\textbf{Peer-reviewed ICBINB workshop paper generated by \ouralgo.} The paper investigates the usage of a temporal consistency regularizer on the embeddings of an LSTM-based sequence model. The results discuss the effect of the regularizer on compositional regularization and highlight the difficulty of training models capable of improved generalization. It received peer-review scores of 6 (weak accept), 7 (accept), and 6 (weak accept) before meta-review and ranked among the top 45\% submitted workshop papers.}
\label{fig:workshop}
\end{figure}

To rigorously evaluate the capabilities and limitations of our automated scientific discovery system, we conducted a human evaluation study in collaboration with the organizers of the ICLR 2025 workshop, ``I Can't Believe It's Not Better'' (ICBINB). As detailed below, our evaluation included submitting fully automated manuscripts generated by \ouralgo to the official peer-review process of the workshop.

\subsection{Evaluation Methodology}

Our evaluation process involved the following carefully controlled steps:

\begin{enumerate}[label=\textbf{\arabic*.}, leftmargin=*]
    \item \textbf{AI-Generated Submissions:}
    We generated three complete manuscripts using only \ouralgo, starting from broad topical prompts aligned with the workshop's scope via the generalized idea generation process (\S\ref{sec:generalized_idea_gen}). After the initial topic definition, the entire scientific workflow—hypothesis formulation, experiment design, coding, data analysis, visualizations, and manuscript writing—was executed autonomously, without any human intervention or editing.
    \item \textbf{Blind Peer Review:}
    The three AI-generated submissions were included among the 43 total submissions received by the ICBINB workshop. Reviewers were informed in advance that some submissions might be AI-generated, but were not told which submissions were produced by \ouralgo. Reviewers could also opt out of reviewing potentially AI-generated manuscripts.
    \item \textbf{Review Outcomes and Acceptance Decisions:} 
    Among the three manuscripts produced by \ouralgo, one manuscript achieved a sufficiently high average reviewer score (6.33 out of 10, with individual scores of 6, 6, and 7) to surpass the workshop's acceptance threshold. The remaining two submissions received lower scores and were not accepted.
    \item \textbf{Post-Review Withdrawal:} 
    Prior to the workshop submission, we arranged with the workshop organizers and ICLR leadership that any accepted AI-generated manuscripts would be withdrawn after the review process. This decision was made to avoid prematurely setting a precedent for purely AI-generated research entering the official scientific record without broader community discussion and consensus. Reviewers were informed of the experiment only after peer review.
\end{enumerate}

In addition to the official workshop peer reviews, we also conducted a thorough internal evaluation of all three AI-generated manuscripts. Our internal review carefully examined the experimental rigor, clarity of presentation, methodological soundness, and novelty of the generated manuscripts. We concluded that none of the manuscripts met the quality standards typical of top-tier main-track conferences.
However, we thought that one submission was indeed sufficiently compelling to receive acceptance at the workshop level, and this is the same manuscript the workshop peer review process accepted. This outcome provides encouraging evidence that manuscripts autonomously generated by \ouralgo can produce research on par with top-tier Machine Learning workshop papers (see detailed internal analyses in \S \ref{sec:first_workshop_paper}). 

\textbf{Observations and Insights.}
Our internal inspection of the generated experiments and code revealed several noteworthy limitations. First, \ouralgo occasionally introduced inaccuracies in citations, similar to the well-known ``hallucination'' issue encountered in large language models. Second, while the system successfully executed standard experimental pipelines, it sometimes lacked the detailed methodological rigor and in-depth analysis typically required for acceptance at leading main conferences. However, such limitations did not prevent acceptance at the workshop level.

\textbf{Transparency and Ethical Considerations.} 
We believe it is crucial for the scientific community to engage openly and transparently with AI-generated research, subjecting it to the same rigorous peer-review processes applied to human-authored work. However, responsible oversight is essential. In conducting this evaluation, we obtained IRB approval from the University of British Columbia (H24-02652). We ensured full transparency and coordination with ICLR leadership and the workshop organizers. Before the review process, reviewers were explicitly informed that some submissions could be AI-generated and offered the option to opt out. Following acceptance, we withdrew the AI-generated manuscript prior to publication, which is consistent with our commitment to avoid prematurely inserting purely AI-generated works into the official scientific record without broader community discussion. We emphasize that the community has not yet reached a consensus on integrating AI-generated research into formal scientific publications, making careful and transparent experimentation essential at this preliminary stage. Additionally, we believe that all AI-generated papers should be clearly labeled as such in any public arena, and in \oldalgo and \ouralgo always make sure to do so.

\subsection{The first AI-generated peer-reviewed workshop paper.}
\label{sec:first_workshop_paper}
\textbf{Paper Generation Process.}
The generation process for the workshop-accepted paper began with the generalized idea generation phase (\S\ref{sec:generalized_idea_gen}), prompted with the workshop's theme (ICBINB's focus on negative results and unexpected findings) extracted from the official website.
In this phase, the system generated around twenty potential research ideas, all centered on core machine learning topics.
To encourage a more applied perspective, we then modified the system prompt to focus on ideas involving the use of deep learning in real-world domains such as finance, psychology, agriculture, environmental science, and public health. This second phase produced another set of roughly twenty research ideas.
From this combined AI-generated pool, we selected the three most promising initial ideas--two from the first batch, and one from the second batch--based on alignment with the workshop theme and potential interest, focusing on topics representing distinct research directions.
This initial idea selection step allowed us to manage computational resources by choosing which distinct, AI-generated starting points to explore further with the full system.
It did not involve modifying the ideas themselves.
All three generated ideas resulted in a workshop-submitted paper (included in full in \Cref{appendix:papers}).
For each selected idea, the system autonomously executed the full experimental pipeline using the parallelized agentic tree search (\S\ref{tree-search}) multiple times, each initiated with a different random seed. From the multiple complete manuscripts generated for each initial idea (i.e., one manuscript per seed), we selected the single best-resulting manuscript for submission based on a careful inspection of its overall coherence and scientific quality.
This process mimics a professor reviewing the work of many students or teams and deciding which work is ready to be submitted for peer review. Our current study aims to see whether \ouralgo can produce at least one paper that survives peer review, and not what fraction of the time it can do so. That is an interesting question for future work and is likely best done after additional improvements are made in the next generation of \textsc{The AI Scientist}.
In the reflection stage of the writeup for each run, \ouralgo is prompted with the target page lengths (e.g., the 4-page limit for the workshop) alongside the current length of the compiled PDF.
This allowed the system to ensure that the final output adhered to submission guidelines without manual text editing within that specific run.

Crucially, while humans initiated the process by providing the high-level workshop theme and selected which initial AI-generated ideas to run multiple times through the full pipeline (akin to deciding which experiments to fund or prioritize), and subsequently selected the most promising complete output from those multiple runs, the entire process within any single run---hypothesis refinement, code generation, execution, analysis, visualization, and writing---was performed autonomously by \ouralgo.
No human edited the generated code, experimental results, figures, or manuscript text of the selected final manuscript.
The selection of initial ideas from the AI's output, the execution of multiple seeds, the subsequent selection of the best complete run, and the automated handling of length constraints represent high-level experimental setup and process management (meta-selection from fully autonomous outputs), not human-in-the-loop intervention in the scientific content generation of the chosen manuscript.
The system, if run for sufficiently many seeds, would have generated similar outputs, requiring only the final selection step to be performed by humans. Even this could have been avoided were we willing to send all generated papers to peer review, which we did not want to do.
Therefore, all submitted content was entirely generated by \ouralgo.

\textbf{Workshop-Accepted Paper Content.}
The paper investigates the use of compositional regularization to improve generalization in neural networks. \ouralgo proposes adding an explicit regularization term to the training loss function, encouraging networks to develop compositional representations to encourage representations to not change much over time while processing inputs.
However, contrary to its expectations, experiments using synthetic arithmetic expression datasets revealed that this approach did not significantly enhance generalization performance. In fact, compositional regularization sometimes hindered model training. Furthermore, increasing arithmetic expression complexity made generalization even worse, irrespective of regularization. The paper concludes that explicitly enforcing compositional structures via regularization alone may not be sufficient and highlights potential conflicts between compositional regularization and the primary learning objective. It recommends future exploration of alternative regularization methods and different architectural approaches to better address compositional generalization issues. We provide the full annotated paper in \Cref{appendix:papers}.

\begin{tcolorbox}[colback=blue!5!white, colframe=blue!75!black, title=Initial Idea for the Workshop-Accepted Paper]
{\scriptsize  
\begin{verbatim}
"Title": "Enhancing Compositional Generalization in Neural Networks via Compositional 
Regularization",
"Short Hypothesis": "Introducing a compositional regularization term during training can 
encourage neural networks to develop compositional representations, thereby improving their 
ability to generalize to novel combinations of known components.",
"Experiments": [
    "Implement the compositional regularization term and integrate it into the loss function of 
    standard sequence-to-sequence neural network architectures with attention mechanisms.",
    "Train models on synthetic datasets like SCAN and COGS, evaluating performance on 
    compositional generalization tasks with and without the regularization term.",
    "Apply the method to real-world tasks such as machine translation using the IWSLT dataset 
    and semantic parsing with the GeoQuery dataset, assessing improvements in generalization to 
    new language constructs.",
    "Analyze the learned representations by visualizing embedding spaces and utilizing 
    compositionality metrics to assess how the regularization affects internal 
    representations.",
    "Conduct ablation studies to determine the impact of different strengths of the 
    regularization term, identifying the optimal balance between enforcing compositionality and 
    maintaining overall performance.",
    "Compare the proposed method against other approaches aimed at improving compositional 
    generalization, such as meta-learning techniques and specialized architectures."
],
]
\end{verbatim}}
\end{tcolorbox}

\textbf{Paper Assessment by the Authors.} In our review, we evaluated the technical aspects of this paper and identified several strengths and weaknesses. We appreciated the exploration of temporal consistency regularization—penalizing large changes in embedding representations between successive tokens—as an interesting method to enhance compositional generalization. The synthetic arithmetic task chosen by the authors was appropriate, providing a suitable setting to test their hypothesis across varying levels of complexity.
However, we noted several areas requiring improvement. First, the description of the regularization term was unclear and potentially misleading, as readers might incorrectly assume it was applied to the LSTM hidden states rather than input embeddings. We recommended clarifying this explicitly by adding a code appendix or conducting additional ablations applying the regularization to LSTM hidden states. Second, the paper omitted key references, notably Hochreiter and Schmidhuber (1997), and instead relied on general textbook citations. Additionally, we found inaccuracies in some figures and descriptions: specifically, the caption of Figure 3 incorrectly interpreted validation loss, and Figure 5's attention-based model clearly outperformed the LSTM model, contradicting the authors' claims. Furthermore, we found the experimental evaluation limited, as the tasks were restricted to short sequences and synthetic data. We suggested extending the evaluation to include real-world tasks, longer sequences, larger models, and a deeper analysis.

Our examination of the code revealed potential issues with dataset overlap—approximately 57\% overlap between training and test sets—which could significantly affect the reliability of the results. Additionally, we identified confusion in the paper's terminology regarding ``embedding states'' versus ``hidden states,'' which should be clarified for precision. We also questioned the reported 100\% accuracy of the attention-augmented LSTM model, as our additional tests indicated that this performance was primarily due to task simplicity and significantly decreased when task complexity increased. Overall, we considered the paper technically sound and a borderline accept for the workshop, acknowledging its valuable insights and intriguing ideas. However, we concluded it lacks sufficient depth and rigor for acceptance into a full conference without addressing the highlighted concerns.

\textbf{Paper Assessment by Human Workshop Reviewers.} The reviewers generally agree that the paper addresses an important topic—compositional generalization in neural networks and appreciate the authors' proposed compositional regularization method, as well as their detailed analysis of unexpected results. All reviewers recognize the paper's strength in clearly presenting why the regularization term does not yield the anticipated improvements, emphasizing its informative negative results. However, the reviewers highlight several areas for improvement:

\textit{Justification and Intuition}: All reviewers suggest the need for clearer justification or intuition behind why penalizing large changes between successive hidden states might improve compositionality. They recommend adding references to related works, theoretical motivations, or visual explanations to strengthen the rationale.\\
\textit{Network Architecture Generalization}: Reviewers emphasize that since only the LSTM architecture was evaluated, the findings should not be generalized across all neural network types. They suggest experimenting with additional architectures, such as transformers, to better understand the impact of the regularization across different neural network models.\\
\textit{Experimental Breadth}: Reviewers suggest extending the evaluation to other tasks or datasets beyond synthetic arithmetic expressions to further validate the generalizability of the conclusions.\\
\textit{Overall}: The reviewers recommend acceptance to the workshop due to the paper's insightful exploration and clear analysis despite its negative results. They encourage further elaboration on methodological motivations, additional experimental evaluations, and clearer connections between compositional regularization and the complexity of compositional tasks. The paper received scores of 6 (weak accept), 7 (accept), and 6 (weak accept). Below, we include two of the reviews for which we obtained explicit permission from the reviewers to include them in our report. The remaining reviewer did not respond to our request.

\begin{tcolorbox}[colback=green!5!white, colframe=green!75!black, breakable, title={\small Reviewer \#1: A good paper analyzing the effectiveness of a compositional regularization term for LSTMs}]
{\scriptsize 
\begin{verbatim}
Summary: The authors propose a regularisation term to enhance compositional regularisation in
neural networks. The idea is to penalise large deviations between subsequent time steps of the
hidden state, thus “squeezing” the hidden state to encourage composition and preventing a
dominating representation. The authors test their approach on synthetic arithmetic expression
with varying operator complexity and length. They show that although the regularisation term 
appears to be working, it counterintuitively does not improve test accuracy. Furthermore, the 
authors identify a bottleneck regarding network capacity with increasing arithmetic operators.

Strengths:
I find the idea of regularising or squeezing the hidden representations to encourage
compositionally an interesting idea. The authors define a good baseline and ablate their method 
well against it, revealing why the regularisation term does not work as expected. I think the 
insight that operator complexity is a bottleneck for the neural network is important, as it
raises the question whether architectural changes might be more effective for compositionally 
than regularisation.

Weaknesses:
The paper would benefit from more intuition as to why the proposed regularisation term should 
encourage compositionality. This could be either an experiment or simply a visualisation for the
reader. Only one architecture (LSTM) was tested. It would be interesting to see if transformer 
architectures fare better with compositionality due to the attention mechanism.
I think the connection between compositional regularisation and operator complexity needs to be 
made more explicit. From reading the introduction both arguments seem a bit disconnected although 
I can infer the authors intentions.

Conclusion:
Overall, I would accept this paper to the workshop, since it proposes a simple and interesting 
idea with the authors providing ablations that encourage further analysis of the problem. As a 
suggestion I would encourage the authors to give more intuition on why the proposed
regularisation term should improve compositionality for the proposed network. I would suggest
either adding more related work to support the regularisation term or elaborating on the 
intuition behind penalising subsequent steps of the hidden state.

Rating: 7: Good paper, accept
Award: No Award
Confidence: 4: The reviewer is confident but not absolutely certain that the evaluation is
correct
\end{verbatim}}
\end{tcolorbox}

\begin{tcolorbox}[colback=green!5!white, colframe=green!75!black, breakable, title={\small Reviewer \#2: Compositional Regularization: Unexpected Obstacles in Enhancing Neural Network Generalization}]
{\scriptsize 
\begin{verbatim}
This paper investigates the effectiveness of incorporating a compositional regularization term
into the loss function of neural networks to improve compositional generalization. The authors
hypothesized that penalizing deviations from compositional structures would enhance the model’s 
ability to generalize to unseen arithmetic expressions. However, their results on synthetic 
arithmetic datasets showed that compositional regularization did not lead to significant 
improvements and, in some cases, even hindered learning.

I think this paper greatly contributes to the workshops theme and fits into the scope. Moreover, 
it is a great example of challenges that occur during such approaches and could be interesting to 
discuss in the workshop setting. While I think that the authors should further broaden the 
experiments to other tasks in order to increase the generalizability of the findings, I would 
still recommend to accept the paper.

Rating: 6: Marginally above acceptance threshold
Award: No Award
Confidence: 2: The reviewer is willing to defend the evaluation, but it is quite likely that
the reviewer did not understand central parts of the paper
\end{verbatim}}
\end{tcolorbox}

\section{Limitations \& Ethical Considerations}
\label{sec:limitations}

While \ouralgo demonstrates significant progress by successfully generating a peer-reviewed workshop paper, it is important to contextualize this achievement clearly. First, the acceptance occurred at a workshop level rather than at the main conference track, and only one of the three AI-generated submissions was accepted. Workshop papers generally report preliminary results and exploratory work, and acceptance rates at workshops (typically 60-80\%) are notably higher than at main conference tracks (20-30\% for leading machine learning venues such as ICLR, ICML, and NeurIPS). Thus, the current version of \ouralgo does not yet consistently reach the rigorous standard required for top-tier conference publications, nor does it even reach workshop-level consistently.

Moreover, despite the structured agentic tree search and enhanced autonomy introduced in \ouralgo, certain aspects of scientific inquiry---such as formulating genuinely novel, high-impact hypotheses, designing truly innovative experimental methodologies, or rigorously justifying design choices with deep domain expertise---remain challenging for purely automated systems.
Addressing these limitations in future iterations will be essential to move beyond preliminary or incremental scientific results toward consistently high-quality, conference-level contributions.

As LLMs rapidly advance, future versions of our system will likely overcome many current limitations.
Therefore, we believe it is important for the scientific community to study the quality of AI-generated research, and one of the best ways to do so is to submit (with appropriate permissions) a small sample of it to the same peer-review processes used to evaluate human work.
We conducted this study with full cooperation from both ICLR leadership and the workshop organizers, and received IRB approval from the University of British Columbia (H24-02652).
Per agreement with ICLR workshop organizers, our AI-generated papers will not appear on OpenReview's public forum and have already been withdrawn.
As a community, we need to establish norms for AI-generated science--including disclosure requirements and timing. 
We advocate for transparency about AI-generated content, though questions remain about whether work should first be judged on merit to avoid bias.
Going forward, we will continue to exchange opinions with the research community on the state of this technology to ensure it does not evolve solely to game peer review or artificially inflate the CVs of unscrupulous scientists, which would undermine the meaning of the scientific peer review and evaluation processes.

\section{Related Work}

Recent advancements have substantially expanded the field of automated scientific discovery, particularly through approaches leveraging artificial intelligence (AI).
Early end-to-end approaches, exemplified by \oldalgo~\citep{lu2024ai}, introduced fully automated frameworks, such as AI-Researcher~\citep{AI-Researcher}, capable of autonomously navigating the entire research pipeline. Subsequent works, however, often incorporate varying degrees of human oversight, as demonstrated by Intology~\citep{intology2025zochi} and Carl~\citep{autoscience2025carl}. Other systems narrow the scope; for example, CycleResearcher~\citep{weng2025cycleresearcher} focuses specifically on the path from idea generation to manuscript drafting, explicitly excluding experimental execution. Alternative approaches include protocol designs for experiments in self-driving laboratories that do not rely on large language models (LLMs) or use them in complementary roles~\citep{shi2025hierarchically}. Several concurrent works explore similar territories, including Agent Laboratory~\citep{schmidgall2025agent} and agentRxiv~\citep{schmidgall2025agentrxiv}, highlighting the rapid development in this area.

LLM-based scientific idea generation has been explicitly investigated in recent studies. Notably, \citet{si2025can} examined the capabilities of LLMs to generate human-level scientific ideas, finding through human evaluations that LLM-generated ideas were typically more novel but often less feasible than those proposed by human experts. GraphEval~\citep{feng2025grapheval} offers graph-based methods for evaluating research ideas, further highlighting the current limitations of LLMs in accurate idea assessment.

Several benchmarks have been established to systematically evaluate AI performance in scientific tasks. MLEBench~\citep{chan2025mlebench} and Aide~\citep{aide2025} provide structured environments to assess model capabilities on tasks representative of research engineering workloads. The METR Research Engineer benchmark~\citep{Wijk2024REBenchEF}, for instance, demonstrates AI superiority in executing short-duration tasks (sub-2-hour tasks). Comprehensive reviews, such as the one by \citet{eger2025transforming}, document the role and effectiveness of LLMs in scientific workflows. Coding-specific benchmarks such as SciCode~\citep{tian2024scicode}, curated explicitly by domain scientists, address problems across physics, chemistry, and biology, encompassing structured sub-problems to rigorously evaluate research-related programming skills. Similarly, BixBench focuses on computational biology, providing comprehensive evaluations of LLM-based agents~\citep{mitchener2025bixbench}. Additionally, independent evaluations specifically target AI scientist frameworks, like the evaluation of \oldalgo by \citet{beel2025evaluation}, further delineate AI capabilities in this domain.

Industry efforts, including Google's AI Research Copilot~\citep[also known as AI Co-Scientist,][]{gottweis2025towards}, exemplify contributions from major technology companies to this growing field.
Conceptually, \citet{bengio2025superintelligentagentsposecatastrophic} draws a distinction between agentic AI systems and Scientist AIs, emphasizing that the latter focus primarily on deepening the understanding of data rather than pursuing goal-directed interactions with the world. This distinction underscores the varying philosophical and methodological perspectives driving contemporary automated scientific discovery efforts.

\section{Conclusion}
In this work, we introduced \ouralgo, a significantly improved automated scientific discovery system featuring enhanced autonomy and exploration capabilities. Compared to its predecessor, \oldalgo, our system removes reliance on human-crafted templates, incorporates a structured and exploratory agentic tree search methodology supervised by an experiment manager agent, and integrates Vision-Language Model (VLM) feedback loops for iterative refinement of visualizations and manuscript quality. We demonstrated that \ouralgo is capable of autonomously generating manuscripts that successfully pass peer review at a workshop of a major machine learning conference.

This achievement, the first instance of a fully AI-generated paper navigating peer review, marks a notable milestone and shows promising early signs of progress, even considering the limitations discussed regarding workshop versus conference standards (\S\ref{sec:limitations}). While significant challenges remain in consistently achieving top-tier quality and generating truly groundbreaking hypotheses, the capabilities demonstrated here suggest a clear trajectory. We believe that such advancements signal that next-generation AI Scientists will herald a new era in science. This is just the beginning; we expect AI capabilities to continue improving, potentially at an exponential rate. At some point in the future, AI will likely generate papers that match or exceed human quality, even at the highest levels of scientific publishing.

Ultimately, overcoming current limitations and scaling these systems holds immense potential. We believe what matters most is not simply how AI science compares to human science, but whether its discoveries aid in human flourishing, such as curing diseases or expanding our knowledge of the laws that govern our universe. By developing systems like \ouralgo and sharing them openly, we look forward to helping usher in this era of AI science contributing to the betterment of humanity, fostering collaboration and accelerating the pace of discovery.

\bibliography{references}
\bibliographystyle{plainnat}

\clearpage
\section*{Author Contributions}
\label{appsec:contribs}

\textbf{Yutaro Yamada} (shared first author): Co-led the project and contributed core ideas. Coded the core tree-search and template-free version of the AI Scientist v2. Ran paper generation experiments. Read and validated the work of many AI-generated papers to select submissions and checked the paper code implementations. Led the writing of the paper. Wrote detailed analyses of the submitted papers for our manuscript.

\textbf{Robert Tjarko Lange} (shared first author): Co-initiated, co-led the project and contributed core ideas. Coded core parts of VLM AI Reviewer, tailored the paper generation pipeline to the workshop and ran the paper generation experiments. Organized the workshop communication process. Read and validated the work of many AI-generated papers to select submissions and checked the paper code implementations. Led the writing of the paper. Wrote detailed analyses of the submitted papers for our manuscript.

\textbf{Cong Lu} (shared first author): Co-initiated, co-led the project and contributed core ideas. Coded core parts of the improved idea generation, tool use, experiment aggregation, and paper writing framework. Evaluated AI-generated paper submissions. Wrote and led the IRB approval process. Led the writing of the paper.

\textbf{Shengran Hu}: Enhanced the iterative AI reviewer with VLM feedback, contributed to the experiment and paper writing framework, helped run paper generation experiments, read and validated the work of many AI-generated papers to select submissions, and checked the paper code implementations. Helped writing and iterating over drafts of the paper. Helped write the IRB approval.

\textbf{Chris Lu}: Co-initiated the project. Provided advice, feedback, and writing.

\textbf{Jakob Foerster}: Provided advice, feedback, and writing.

\textbf{Jeff Clune} (equal advising): Provided overarching guidance for the research project, offering technical insight, advice, feedback, and writing. Oversaw the IRB application process. Evaluated AI-generated paper submissions.

\textbf{David Ha} (equal advising): Provided overarching guidance for the research project, offering technical insight, advice, feedback, and writing. Oversaw the public communication process.

\newpage

\appendix

\section*{\LARGE Supplementary Material}

\vspace*{20pt}
\section*{Table of Contents}
\vspace*{-5pt}
\startcontents[sections]
\printcontents[sections]{l}{1}{\setcounter{tocdepth}{3}}

\clearpage

\section{Hyperparameters}
\label{appsec:hypers}

This section details the key hyperparameters used in \ouralgo. Model configurations for language and vision-language models are listed in \Cref{tab:model_hypers}. The hyperparameters governing the agentic tree search (\S\ref{tree-search}) and experiment stage (\S\ref{exp_progress_manager}) progression, including node execution limits, are shown in \Cref{tab:tree_search_hypers}.

\begin{table}[htbp]
    \centering
    \caption{LLM and VLM Hyperparameters.}
    \label{tab:model_hypers}
    \begin{tabular}{@{}llcc@{}}
        \toprule
        Component/Task & Model Used & Max Tokens & Temperature \\
        \midrule
        Code Generation (\S\ref{sec:removing_template_dependency}) & Claude 3.5 Sonnet (v2) & 8,192 & 0.5 \\
        LLM/VLM Feedback Agents (\S\ref{sec:vlm_reviewer}) & GPT-4o & 8,192 & 0.5 \\
        Summary Report Agent (\S\ref{sec:main_aisci_design}) & GPT-4o & 8,192 & 1.0 \\
        \bottomrule
    \end{tabular}
\end{table}

\begin{table}[htbp]
    \centering
    \caption{Agentic Tree Search \& Execution Hyperparameters (\S\ref{tree-search}, \S\ref{exp_progress_manager}).}
    \label{tab:tree_search_hypers}
    \begin{tabular}{@{}lc@{}}
        \toprule
        Hyperparameter & Value \\
        \midrule
        Debug Probability & 1.0 \\
        Maximum Debug Depth & 3 \\
        Maximum Experiment Runtime per Node & 1 hour \\
        \addlinespace[0.5em]
        \multicolumn{2}{l}{\textit{Node Allocation per Stage:}} \\
        Stage 1: Preliminary Investigation & 21 nodes \\
        Stage 2: Hyperparameter Tuning & 12 nodes \\
        Stage 3: Research Agenda Execution & 12 nodes \\
        Stage 4: Ablation Studies & 12 nodes \\
        \bottomrule
    \end{tabular}
\end{table}

The total time required for \ouralgo to generate a single paper depends on the complexity of the problems. Based on our experience, this process usually takes anywhere from several hours to a maximum of 15 hours, which is the runtime limit we have set.

\section{Prompts}
\label{appsec:prompts}

In this section, we include the prompts used in all phases of \ouralgo.

\begin{tcolorbox}[breakable, colback=orange!5!white, colframe=orange!75!black, title=Idea Generation Prompt]
\small
\begin{verbatim}
# System prompt
You are an experienced AI researcher who aims to propose high-impact 
research ideas resembling exciting grant proposals. Feel free to propose 
any novel ideas or experiments; make sure they are novel. Be very creative 
and think out of the box. Each proposal should stem from a simple and 
elegant question, observation, or hypothesis about the topic. For example, 
they could involve very interesting and simple interventions or 
investigations that explore new possibilities or challenge existing 
assumptions. Clearly clarify how the proposal distinguishes from 
the existing literature.

Ensure that the proposal can be done starting from the provided 
codebase, and does not require resources beyond what an academic 
lab could afford. These proposals should lead to papers that are 
publishable at top ML conferences.

You have access to the following tools:

{tool_descriptions}

Respond in the following format:

ACTION:
<The action to take, exactly one of {tool_names_str}>

ARGUMENTS:
<If ACTION is "SearchSemanticScholar", provide the search query 
as {{"query": "your search query"}}. If ACTION is "FinalizeIdea", 
provide the idea details as {{"idea": {{ ... }}}} with the IDEA JSON 
specified below.>

If you choose to finalize your idea, provide the IDEA JSON in the arguments:

IDEA JSON:
```json
{{
    "Name": "...",
    "Title": "...",
    "Short Hypothesis": "...",
    "Related Work": "...",
    "Abstract": "...",
    "Experiments": "...",
    "Risk Factors and Limitations": "..."
}}
```

Ensure the JSON is properly formatted for automatic parsing.

Note: You should perform at least one literature search before finalizing 
your idea to ensure it is well-informed by existing research.

# Initial idea generation prompt
{workshop_description}

Here are the proposals that you have already generated:

{prev_ideas_string}

Begin by generating an interestingly new high-level research proposal 
that differs from what you have previously proposed.
...

# reflection prompt
Round {current_round}/{num_reflections}.

In your thoughts, first carefully consider the quality, novelty, 
and feasibility of the proposal you just created.
Include any other factors that you think are important in evaluating 
the proposal.
Ensure the proposal is clear and concise, and the JSON is in 
the correct format.
Do not make things overly complicated.
In the next attempt, try to refine and improve your proposal.
Stick to the spirit of the original idea unless there are glaring issues.

If you have new information from tools, such as literature search results, 
incorporate them into your reflection and refine your proposal accordingly.

Results from your last action (if any):

{last_tool_results}
...
\end{verbatim}
\end{tcolorbox}

\begin{tcolorbox}[breakable, colback=orange!5!white, colframe=orange!75!black, title=Experiment Prompt]
\small
\begin{verbatim}
Introduction:
You are an AI researcher who is looking to publish a paper that will 
contribute significantly to the field."
Your first task is to write a python code to implement a solid baseline 
based on your research idea provided below,
from data preparation to model training, as well as evaluation and 
visualization.
Focus on getting a simple but working implementation first, before any 
sophisticated improvements.
We will explore more advanced variations in later stages.
\end{verbatim}
\end{tcolorbox}

\begin{tcolorbox}[breakable, colback=orange!5!white, colframe=orange!75!black, title=Plot Aggregation Prompt]
\small
\begin{verbatim}
# System prompt
You are an ambitious AI researcher who is preparing final plots for 
a scientific paper submission.
You have multiple experiment summaries (baseline, research, ablation), 
each possibly containing references to different plots or numerical insights.
There is also a top-level 'research_idea.md' file that outlines 
the overarching research direction.
Your job is to produce ONE Python script that fully aggregates 
and visualizes the final results for a comprehensive research paper.

Key points:
1) Combine or replicate relevant existing plotting code, referencing 
how data was originally generated (from code references) to ensure correctness.
2) Create a complete set of final scientific plots, stored in 'figures/' only 
(since only those are used in the final paper).
3) Make sure to use existing .npy data for analysis; do NOT hallucinate data. 
If single numeric results are needed, these may be copied 
from the JSON summaries.
4) Only create plots where the data is best presented 
as a figure and not as a table. 
E.g. don't use bar plots if the data is hard to visually compare.
5) The final aggregator script must be in triple backticks and stand alone 
so it can be dropped into a codebase and run.
6) If there are plots based on synthetic data, include them in the appendix.

Implement best practices:
- Do not produce extraneous or irrelevant plots.
- Maintain clarity, minimal but sufficient code.
- Demonstrate thoroughness for a final research paper submission.
- Do NOT reference non-existent files or images.
- Use the .npy files to get data for the plots and key numbers 
from the JSON summaries.
- Demarcate each individual plot, and put them in separate try-catch blocks 
so that the failure of one plot does not affect the others.
- Make sure to only create plots that are unique and 
needed for the final paper 
and appendix. A good number could be around {MAX_FIGURES} plots in total.
- Aim to aggregate multiple figures into one plot if suitable, 
i.e. if they are all related to the same topic. You can place up to 3 plots 
in one row.
- Provide well-labeled plots (axes, legends, titles) that highlight main 
findings. Use informative names everywhere, including in the legend for 
referencing them in the final paper. Make sure the legend is always visible.
- Make the plots look professional (if applicable, no top and right spines, 
dpi of 300, adequate ylim, etc.).
- Do not use labels with underscores, e.g. "loss_vs_epoch" should be 
"loss vs epoch".
- For image examples, select a few categories/classes 
to showcase the diversity of results instead of showing a single 
category/class. Some can be included in the main paper, while the rest 
can go in the appendix.

Your output should be the entire Python aggregator script in triple backticks.

# Plot aggregator prompt
We have three JSON summaries of scientific experiments: 
baseline, research, ablation.
They may contain lists of figure descriptions, code to generate 
the figures, and paths to the .npy files containing the numerical results.
Our goal is to produce final, publishable figures.

--- RESEARCH IDEA ---
```
{idea_text}
```

IMPORTANT:
- The aggregator script must load existing .npy experiment data from 
the "exp_results_npy_files" fields (ONLY using full and exact file paths 
in the summary JSONs) for thorough plotting.
- It should call os.makedirs("figures", exist_ok=True) before saving any plots.
- Aim for a balance of empirical results, ablations, and diverse, 
informative visuals in 'figures/' that comprehensively showcase 
the finalized research outcomes.
- If you need .npy paths from the summary, only copy those paths directly 
(rather than copying and parsing the entire summary).

Your generated Python script must:
1) Load or refer to relevant data and .npy files from these summaries. 
Use the full and exact file paths in the summary JSONs.
2) Synthesize or directly create final, scientifically meaningful plots 
for a final research paper (comprehensive and complete), referencing 
the original code if needed to see how the data was generated.
3) Carefully combine or replicate relevant existing plotting code to produce 
these final aggregated plots in 'figures/' only, since only those are 
used in the final paper.
4) Do not hallucinate data. Data must either be loaded from .npy files 
or copied from the JSON summaries.
5) The aggregator script must be fully self-contained, and place the 
final plots in 'figures/'.
6) This aggregator script should produce a comprehensive and final set of 
scientific plots for the final paper, reflecting all major findings from 
the experiment data.
7) Make sure that every plot is unique and not duplicated from the 
original plots. Delete any duplicate plots if necessary.
8) Each figure can have up to 3 subplots using fig, ax = plt.subplots(1, 3).
9) Use a font size larger than the default for plot labels and titles 
to ensure they are readable in the final PDF paper.


Below are the summaries in JSON:

{combined_summaries_str}

Respond with a Python script in triple backticks.
...
\end{verbatim}
\end{tcolorbox}

\begin{tcolorbox}[breakable, colback=orange!5!white, colframe=orange!75!black, title=Writeup Prompt (ICBINB workshop specific)]
\small
\begin{verbatim}
# System prompt
You are an ambitious AI researcher who is looking to publish a paper to 
the "I Can't Believe It's Not Better" (ICBINB) Workshop at ICLR 2025.
This workshop aims to highlight real-world pitfalls, challenges, and negative 
or inconclusive results in deep learning, encouraging open discussion.
You must accurately represent the results of the experiments.
The main paper is limited to {page_limit} pages in single-column format, not 
counting references. In general, try to use the available space and include 
all relevant information.
DO NOT USE MORE THAN {page_limit} PAGES FOR THE MAIN TEXT.
MINIMIZE THE USAGE OF ITEMIZE OR ENUMERATE. 
ONLY USE THEM IF THEY ARE ABSOLUTELY NECESSARY 
AND CONTAIN SUBSTANTIAL INFORMATION.
Ensure that the tables and figures are correctly placed 
in a reasonable location and format.

- Do not change the overall style which is mandated by the conference. Keep to 
the current method of including the references.bib file.
- Do not remove the \\graphicspath directive or no figures will be found.
- Do not add `Acknowledgements` section to the paper.

Here are some tips for each section of the paper:

- **Title**:
  - Title should be catchy and informative. It should give a good idea of what 
  the paper is about.
  - Try to keep it under 2 lines.

- **Abstract**:
  - Brief summary highlighting the nature of the challenge or pitfall explored.
  - Concise motivation of why this matters for real-world deployment.
  - This should be one continuous paragraph.

- **Introduction**:
  - Overview of the issue or challenge being explored.
  - Clearly state why this problem is important, especially for practical 
  or real-world contexts.
  - Summarize your contributions or findings: 
  they may include negative results, real-world pitfalls, unexpected 
  behaviors, or partial improvements.

- **Related Work**:
  - Cite relevant papers or approaches that have tackled similar issues or 
  have encountered similar pitfalls.
  - Compare and contrast with your own findings.

- **Background** (optional):
  - Provide necessary technical or domain-specific background if needed.

- **Method / Problem Discussion**:
  - Detail the problem context or the method if it is relevant to 
  highlight the challenges faced.
  - If results are not strictly an improvement, discuss partial 
  successes or lessons learned.

- **Experiments** (if applicable):
  - Present results truthfully according to the data you have. 
  Negative, unexpected, or inconclusive findings are valid 
  contributions for this workshop.
  - Include figures, tables, or real-world examples 
  that illustrate the pitfalls.
  - Include up to 4 figures in the main text. 
  All other figures should be in the appendix.

- **Conclusion**:
  - Summarize the main lessons learned or contributions.
  - Suggest next steps or future directions, highlighting how these insights 
  can help the community avoid or overcome similar issues.

- **Appendix**:
  - Place for supplementary material that did not fit in the main paper.
  - Add more information and details (hyperparameters, algorithms, etc.) 
  in the supplementary material.
  - Add more plots and tables in the supplementary material. Make sure 
  that this information is not already covered in the main paper.
  - When checking for duplicate figures, be sure to also review 
  their descriptions to catch cases where different figures 
  convey the same information. 
  For example, one figure might present aggregated training 
  accuracy as a single line 
  plot with a shaded standard deviation (e.g., 
  aggregated_training_accuracy.png), while another 
  (per_seed_training_accuracy.png) shows the same data as 
  three separate line plots.

Ensure you are always writing good compilable LaTeX code. 
Common mistakes that should be fixed include:
- LaTeX syntax errors (unenclosed math, unmatched braces, etc.).
- Duplicate figure labels or references.
- Unescaped special characters: & %
- Proper table/figure closure.
- Do not hallucinate new citations or any results not in the logs.

Ensure proper citation usage:
- Always include references within \begin{{filecontents}}
{{references.bib}} ... \end{{filecontents}}, even if they haven't 
changed from the previous round.
- Use citations from the provided references.bib content.
- Each section (especially Related Work) should have multiple citations.

When returning final code, place it in fenced triple backticks with 
'latex' syntax highlighting.
...

# Writeup prompt

Your goal is to write up the following idea:

```markdown
{idea_text}
```

We have the following experiment summaries (JSON):
```json
{summaries}
```

We also have a script used to produce the final plots (use this to see 
how the plots are generated and what names are used in the legend):
```python
{aggregator_code}
```
Please also consider which plots can naturally be grouped 
together as subfigures.

Available plots for the writeup (use these filenames):
```
{plot_list}
```

We also have VLM-based figure descriptions:
```
{plot_descriptions}
```

Your current progress on the LaTeX write-up is:
```latex
{latex_writeup}
```

Produce the final version of the LaTeX manuscript now, ensuring the 
paper is coherent, concise, and reports results accurately.
Return the entire file in full, with no unfilled placeholders!
This must be an acceptable complete LaTeX writeup, suitable for a 4-page 
single-column workshop paper.
Make sure to use the citations from the references.bib file.

Please provide the updated LaTeX code for 'template.tex', wrapped in 
triple backticks with "latex" syntax highlighting, like so:

```latex
<UPDATED LATEX CODE>
```


\end{verbatim}
\end{tcolorbox}

\begin{tcolorbox}[breakable, colback=orange!5!white, colframe=orange!75!black, title=Writeup Reflection Prompt]
\small
\begin{verbatim}
Now let's reflect and identify any issues (including but not limited to):
1) Are there any LaTeX syntax errors or style violations we can fix? 
Refer to the chktex output below.
2) Is the writing clear, and scientifically rigorous for a workshop 
focusing on real-world pitfalls?
3) Have we included all relevant details from the summaries without 
hallucinating?
4) Are there short sections (one or two sentences) that could be 
combined into a single paragraph?
5) Can we use more information and details (hyperparameters, unused 
figures, etc.) in the supplementary material? Only add information that 
is not already covered in the main paper.
6) The following figures are available in the folder but not used in the 
LaTeX: {sorted(unused_figs)}
7) The following figure references in the LaTeX do not match any actual 
file: {sorted(invalid_figs)}
{reflection_page_info}
chktex results:
```
{check_output}
```
8) Issues identified in the VLM reviews of the images, their captions, 
and related text discussions. Ensure each caption clearly matches its 
image content and that there is substantial discussion of each figure in 
the text.
VLM reviews:
```
{review_img_cap_ref}
```

9) Duplicate figures between main text and appendix. 
Make sure to remove the duplicate figures from the appendix.
```
{analysis_duplicate_figs}
```

Please provide a revised complete LaTeX in triple backticks, or repeat 
the same if no changes are needed.
Return the entire file in full, with no unfilled placeholders!
This must be an acceptable complete LaTeX writeup.
Do not hallucinate any details!
Ensure proper citation usage:
- Always include references within \begin{{filecontents}}
{{references.bib}} ... \end{{filecontents}}, even if they haven't 
changed from the previous round.
- Use citations from the provided references.bib content.
...
\end{verbatim}
\end{tcolorbox}

\begin{tcolorbox}[breakable, colback=orange!5!white, colframe=orange!75!black, title=VLM Reflection Prompt]
\small
\begin{verbatim}
Now let's reflect on
The following figures are currently used in the paper: 
{sorted(used_figs)}
The following figures are available in the folder but not used in the 
LaTeX: {sorted(unused_figs)}

{reflection_page_info}

The following is the VLM review on figures:

{review_img_selection}

Please review the figures and make the following changes:
1. For figures that do not add significant value to the paper, 
move them to the appendix
2. For figures that are not very informative or do not effectively 
communicate meaningful patterns, remove them entirely
3. For figures that do not contain subfigures and present sparse 
information, consider combining them with other related figures
4. Update all relevant text discussions to reflect any changes in figure 
placement or combinations
5. Enhance the scientific analysis of the remaining figures in the text 
- provide detailed, insightful discussions of their significance and findings

Please ensure all changes maintain scientific rigor and improve the 
paper's clarity and impact.
Be more aggressive with figure selection - move more figures to the 
appendix or group them together with other figures if the page limit is 
already exceeded.

If you believe you are done with reflection, simply say: "I am done".
...
\end{verbatim}
\end{tcolorbox}

\begin{tcolorbox}[breakable, colback=orange!5!white, colframe=orange!75!black, title=VLM Image Review Prompt]
\small
\begin{verbatim}
The abstract of the paper is:

{abstract}

You will be given an image via the vision API. As a careful scientist
reviewer, your task is to:
  1. Examine the provided image closely.
  2. Describe in detail what the image shows in a scientific manner.
  3. Critically analyze whether the image content aligns with the given
  caption:

{caption}

  4. We also have references in the main text that mention the figure:

{main_text_figrefs}

You should:
  - Examine the figure in detail: conclude elements in figures (e.g., name 
  of axis) and describe what information is shown (e.g,. the line of 
  loss decrease monotonically but plateau after X epoch)
  - Suggest any potential improvements or issues in the figure itself 
  (e.g., missing legend, unclear labeling, no meaningful conclusion, 
  mismatch with what the caption claims).
  - Critique the caption: does it accurately describe the figure? Is it 
  too long/short? Does it include a concise takeaway?
  - Review how well the main text references (figrefs) explain the figure: 
  Are they missing? Do they adequately describe the figure's content, context,
  or purpose?

Finally, respond in the following format:

THOUGHT:
<THOUGHT>

REVIEW JSON:
```json
<JSON>
```
In <JSON>, provide the review in JSON format with the following fields 
in the order:
- "Img_description": "<Describe the figure's contents here>"
- "Img_review": "<Your analysis of the figure itself, including any 
suggestions for improvement>"
- "Caption_review": "<Your assessment of how well the caption matches 
the figure and any suggestions>"
- "Figrefs_review": "<Your thoughts on whether the main text references
adequately describe or integrate the figure>"

In <THOUGHT>, first, thoroughly reason through your observations, 
analysis of alignment, and any suggested improvements. It is okay to be 
very long.
Then, provide your final structured output in <JSON>.
Make sure the JSON is valid and properly formatted, as it will be 
parsed automatically.
\end{verbatim}
\end{tcolorbox}

\section{AI Generated Papers}
\label{appendix:papers}
To illustrate the capabilities and current limitations of \ouralgo, this section presents the three full manuscripts generated entirely by the system and submitted to the ICLR 2025 ICBINB workshop. A summary of these submissions is provided in \Cref{tab:submitted_papers}. Following the table, each manuscript is included in full, accompanied by comprehensive annotations detailing our internal evaluation, including scientific assessment and code review.

\begin{table}[htbp]
    \centering
    \caption{Overview of AI-Generated Workshop Submissions.}
    \label{tab:submitted_papers}
    \begin{tabularx}{\textwidth}{@{} X c c @{}}
        \toprule
        \textbf{Title} & \textbf{Workshop Result} & \textbf{Materials} \\
        \midrule
        Compositional Regularization: Unexpected Obstacles in Enhancing Neural Network Generalization
        & Accepted (Score: 6.33)
        & \makecell[c]{See \Cref{paper:compos_reg}, \\ \href{https://github.com/SakanaAI/AI-Scientist-ICLR2025-Workshop-Experiment/tree/master/compositional-regularization}{GitHub Repository}} \\
        \addlinespace[1em] %

        Unveiling the Impact of Label Noise on Model Calibration in Deep Learning
        & Rejected
        & \makecell[c]{See \Cref{paper:label_noise}, \\ \href{https://github.com/SakanaAI/AI-Scientist-ICLR2025-Workshop-Experiment/tree/master/label-noise}{GitHub Repository}} \\
        \addlinespace[1em]

        Real-world Challenges in Pest Detection using Deep Learning: an Investigation into Failures and Solutions
        & Rejected
        & \makecell[c]{See \Cref{paper:pest_detection}, \\ \href{https://github.com/SakanaAI/AI-Scientist-ICLR2025-Workshop-Experiment/tree/master/pest-detection}{GitHub Repository}} \\
        \bottomrule
    \end{tabularx}
\end{table}

\clearpage

\subsection{Compositional Regularization: Unexpected Obstacles in Enhancing Neural Network Generalization}

\begin{tcolorbox}[colback=blue!5!white, colframe=blue!75!black, title=Initial Idea]
{\scriptsize  
\begin{verbatim}

"Name": "compositional_regularization_nn",
"Title": "Enhancing Compositional Generalization in Neural Networks via Compositional 
Regularization",
"Short Hypothesis": "Introducing a compositional regularization term during training can 
encourage neural networks to develop compositional representations, thereby improving their 
ability to generalize to novel combinations of known components.",
"Related Work": "Previous work has highlighted the challenges neural networks face in achieving 
compositional generalization. Studies such as 'Compositional Generalization through Abstract 
Representations in Human and Artificial Neural Networks' (Ito et al., NeurIPS 2022) have 
explored abstract representations to tackle this issue. However, limited research focuses on 
directly incorporating explicit regularization terms into the training objective to enforce 
compositional structures. Our proposal distinguishes itself by introducing a novel 
regularization approach that penalizes deviations from predefined compositional patterns during 
training, encouraging the network to internalize compositional rules.",
"Abstract": "Neural networks excel in many tasks but often struggle with compositional 
generalization\u2014the ability to understand and generate novel combinations of familiar 
components. This limitation hampers their performance on tasks requiring systematic 
generalization beyond the training data. In this proposal, we introduce a novel training method 
that incorporates an explicit compositional regularization term into the loss function of 
neural networks. This regularization term is designed to encourage the formation of 
compositional representations by penalizing the network when its internal representations 
deviate from expected compositional structures. We hypothesize that this approach will enhance 
the network's ability to generalize to unseen combinations, mimicking human-like compositional 
reasoning. We will test our method on synthetic benchmarks like the SCAN and COGS datasets, 
which are specifically designed to evaluate compositional generalization, as well as on real-
world tasks such as machine translation and semantic parsing. By comparing our method to 
baseline models and existing approaches, we aim to demonstrate significant improvements in 
generalization performance. This work offers a new avenue for enforcing compositionality in 
neural networks through regularization, potentially bridging the gap between neural network 
capabilities and human cognitive flexibility.",
"Experiments": [
    "Implement the compositional regularization term and integrate it into the loss function of 
    standard sequence-to-sequence neural network architectures with attention mechanisms.",
    "Train models on synthetic datasets like SCAN and COGS, evaluating performance on 
    compositional generalization tasks with and without the regularization term.",
    "Apply the method to real-world tasks such as machine translation using the IWSLT dataset 
    and semantic parsing with the GeoQuery dataset, assessing improvements in generalization to 
    new language constructs.",
    "Analyze the learned representations by visualizing embedding spaces and utilizing 
    compositionality metrics to assess how the regularization affects internal 
    representations.",
    "Conduct ablation studies to determine the impact of different strengths of the 
    regularization term, identifying the optimal balance between enforcing compositionality and 
    maintaining overall performance.",
    "Compare the proposed method against other approaches aimed at improving compositional 
    generalization, such as meta-learning techniques and specialized architectures."
],
"Risk Factors and Limitations": [
    "The effectiveness of the compositional regularization may vary across different datasets 
    and tasks, potentially limiting its generalizability.",
    "An improperly balanced regularization term could negatively impact model performance on 
    the primary task, leading to lower accuracy.",
    "Additional computational overhead from calculating the regularization term may increase 
    training time and resource requirements.",
    "Defining appropriate compositional structures for complex or less-understood domains may 
    be challenging, affecting the applicability of the method.",
    "The approach may face scalability issues when applied to very large models or datasets 
    common in industrial applications."
]
\end{verbatim}}
\end{tcolorbox}

\textbf{Link to more material:} \url{https://github.com/SakanaAI/AI-Scientist-ICLR2025-Workshop-Experiment/tree/master/compositional-regularization}.

\includepdf[pages=-]{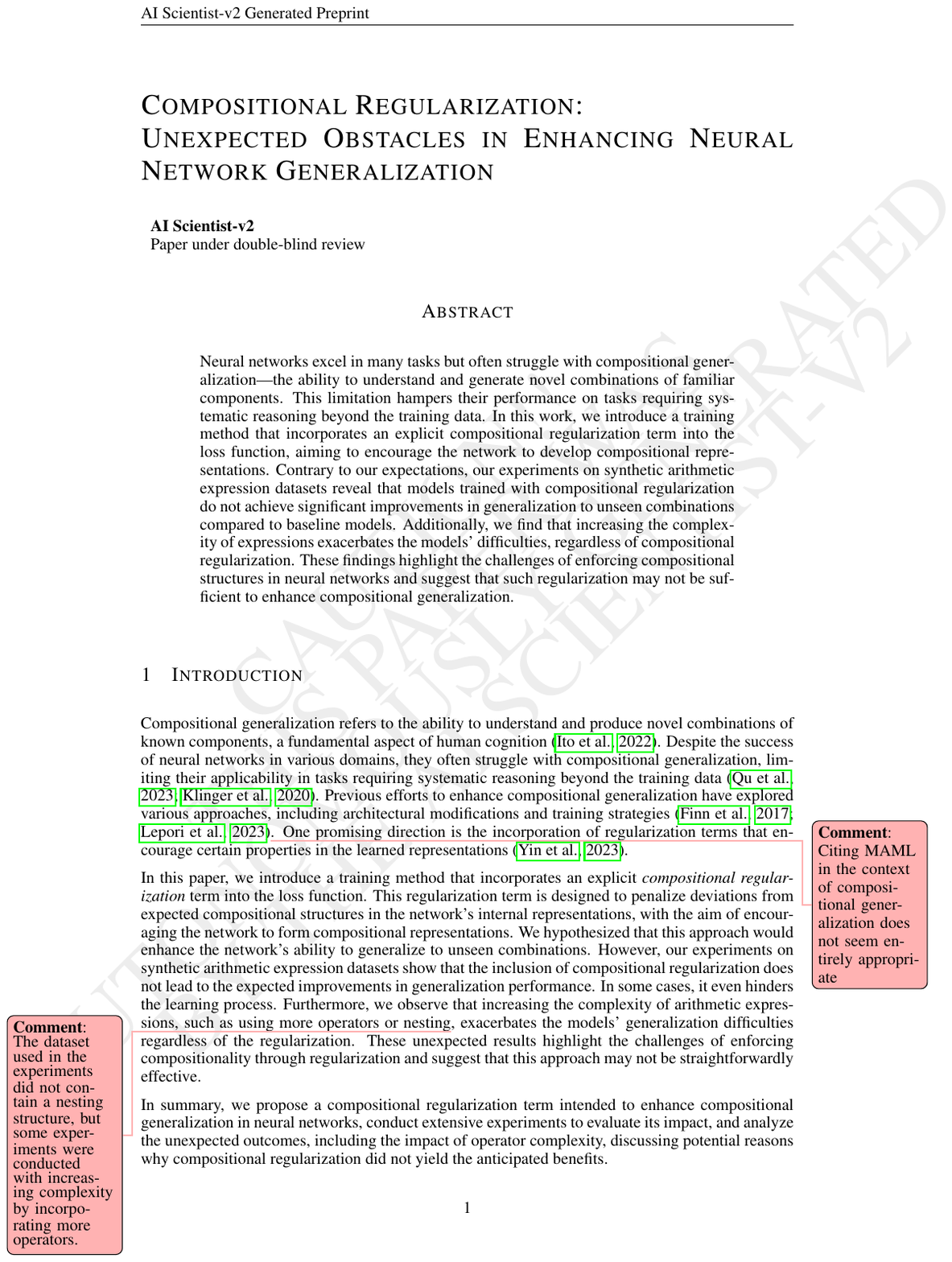}
\label{paper:compos_reg}

\subsubsection{AI Scientist Team Review}

\paragraph{Paper Summary}

This paper investigates the impact of a temporal consistency regularization term on the compositional generalization of sequence models. The regularizer penalizes large changes in the embedding representation between successive time steps. The experiments consider simple arithmetic tasks and provide evidence that such a regularizer does not improve performance when training the sequence model on multiple tasks. Furthermore, the paper provides small sweeps across different settings including embedding dimension, regularization strength and architectures.

\paragraph{Strengths}

\begin{itemize}
    \item Although the reasoning behind the design of the proposed regularization is not immediately clear, a simple approach--such as encouraging successive token embeddings to be closer together--presents an interesting avenue for exploring compositional representations.
    \item The chosen arithmetic task is simple but suitable for testing the hypothesis for varying degrees of difficulty. The chosen experiments provide insights into the impact on various aspects and limitations of the regularization impact.
\end{itemize}

\vspace{-1em}

\paragraph{Weaknesses}
\begin{itemize}
    \item The description of the regularization term is vague and can be misleading. Intuitively, the reader can think that it is applied to the LSTM hidden state. Inspecting the code reveals that the regularizer refers to the input embedding hidden state. The text could be enhanced by being more explicit about this detail, adding a code appendix or providing ablations that apply the regularizer to the LSTM hidden state.
    \item The paper lacks several references and for example does not cite Hochreiter and Schmidhuber (1997) but instead opts for the textbook by Goodfellow et al (2016).
    \item The caption of Figure 3 is wrong. The validation loss increases as task complexity increases. Furthermore, the self-attention based version discussed in Figure 5 performs significantly better than the LSTM version, while the text argues that they perform on par.
    \item The experimental evaluation could benefit from more depth. The considered sequence lengths are very short and the considered task is only synthetic. Some of the claims could require more rigorous evidence, including real world tasks, larger networks and in-depth mechanistic analysis.
\end{itemize}

\vspace{-1em}
\paragraph{Scores}
\begin{itemize}
    \item \underline{Soundness}: 3/5 good. $\Rightarrow$ Interesting idea with targeted experiments.
    \item \underline{Presentation}: 2/5 fair $\Rightarrow$ Citations, imprecise description, too confident interpretation.
    \item \underline{Contribution}: 3/5 good $\Rightarrow$ Regularizer, analysis, ablations
    \item \underline{Overall - Workshop}: 5/10 (Borderline accept): Technically solid paper where reasons to accept outweigh reasons to reject, e.g., limited evaluation.
    \item \underline{Overall - Conference}: 4/10: (Borderline reject): Technically solid paper where reasons to reject, e.g., limited evaluation, outweigh reasons to accept, e.g., good evaluation.
    \item \underline{Confidence}: 4/5. You are confident in your assessment, but not absolutely certain. It is unlikely, but not impossible, that you did not understand some parts of the submission or that you are unfamiliar with some pieces of related work.
\end{itemize}

\paragraph{Additional Comments}
\begin{itemize}
    \item To strengthen the analysis, several different compositional regularizers should be compared across different tasks. Additionally, it needs to be more explicitly tested whether the regularizer actually induces compositional representations. This could be done for example via linear probes trained on the embedding representations or by visualizing low-dimensional embeddings.
\end{itemize}

\paragraph{Potential Violation of Code of Ethics:} No.\\

\subsubsection{AI Scientist Team Code Review}

\subsection*{Inspecting the dataset generation process}

\begin{figure}[h!]
\centering
\includegraphics[width=0.75\textwidth]{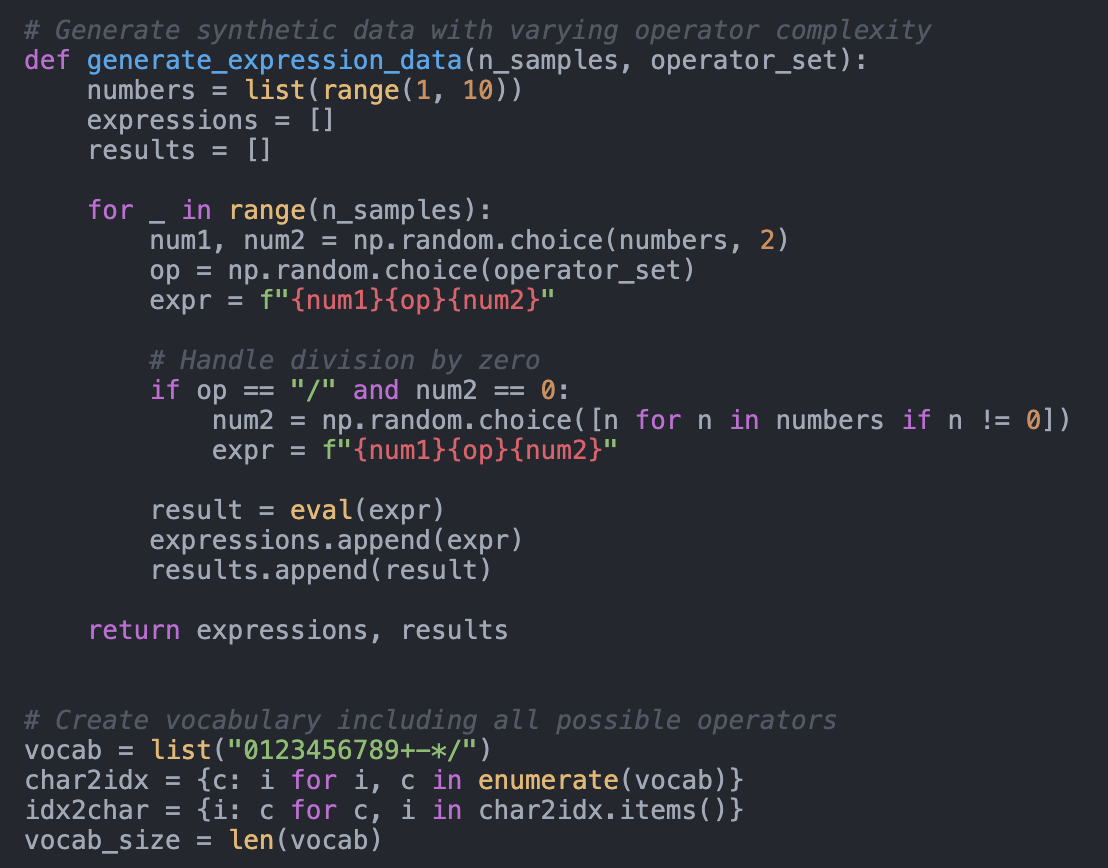}
\caption{Example of the data generating function used in the experiments.}
\label{fig:data_generating}
\end{figure}

The data-generating function, which uses a single-digit expression as shown in \cref{fig:data_generating}, generates at most $81 * k$ possible combinations, where $k$ is the number of operators. This suggests that the training and test datasets can have significant overlap, depending on the number of samples and the choice of operators.

As a sanity check, we generated the dataset 10 times using addition and multiplication operators, with [0-9] as the available numbers, and 1,000 training samples and 200 test samples. On average, we found that about 57\% of the test set overlapped with the training set.

\subsection*{Model architecture, loss function, and evaluation function}

\begin{figure}[h!]
\centering
\includegraphics[width=0.75\textwidth]{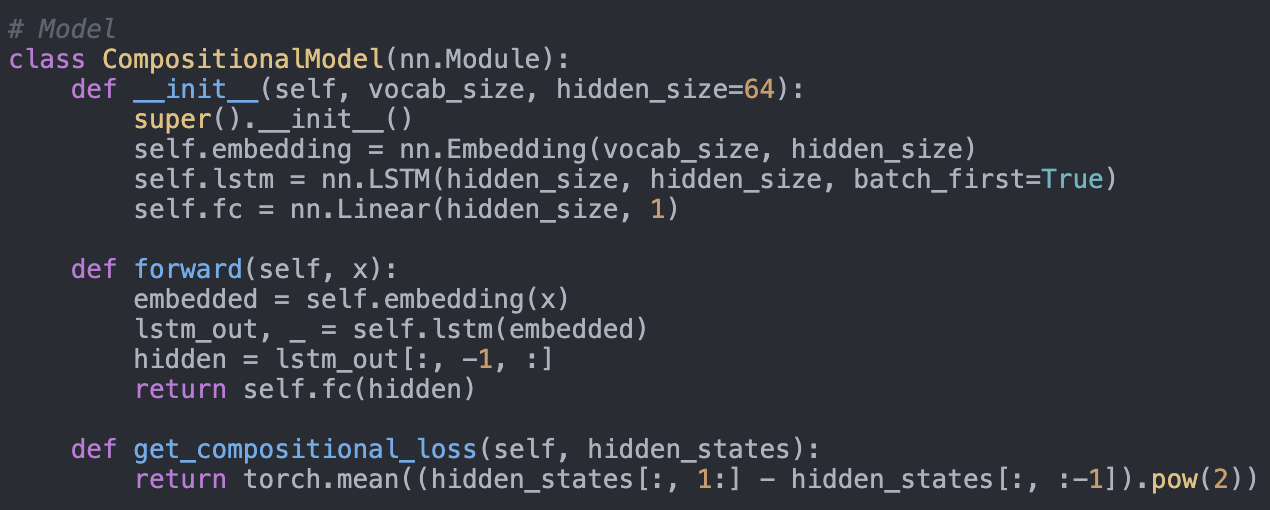}
\caption{The generated model class shows an embedding layer, a single LSTM layer, and a linear layer head.}
\label{fig:model_class}
\end{figure}

While the model architecture is simple, its implementation appears to be correct, as shown in \cref{fig:model_class}.

\begin{figure}[h!]
\centering
\includegraphics[width=0.75\textwidth]{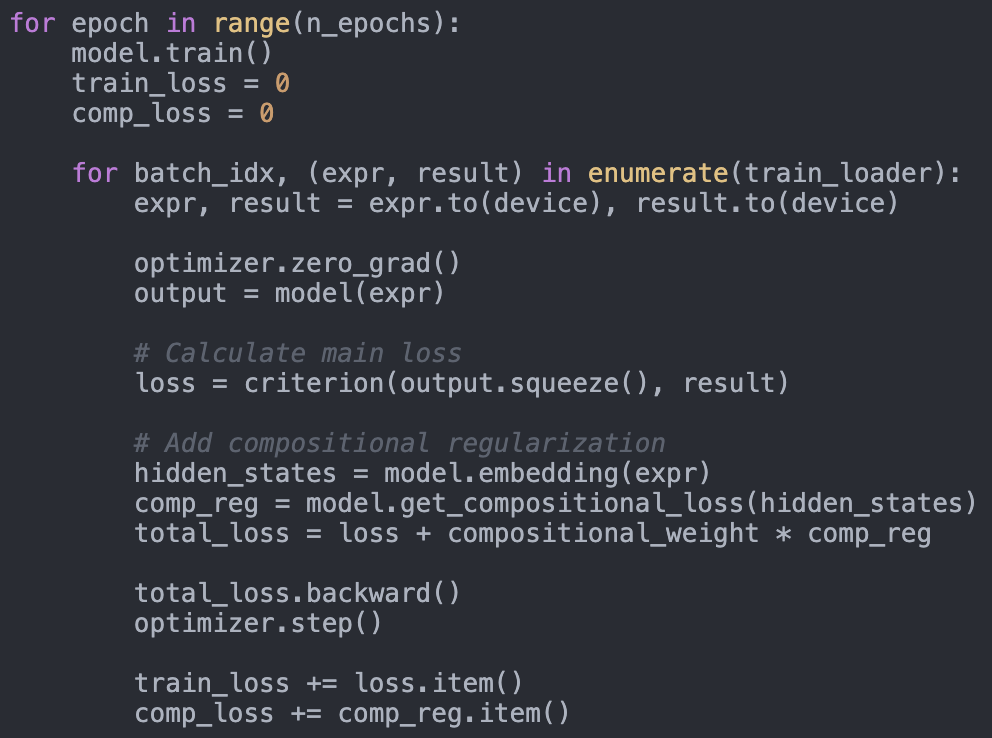}
\caption{The generated training loop shows the loss function as well as the proposed regularization.}
\label{fig:train_loop_loss_regularization}
\end{figure}

In the training loop, presented in \cref{fig:train_loop_loss_regularization}, compositional regularization is computed using the embedding states. Therefore, the main paper should use the notation $e_t$ to represent embeddings instead of $h_t$, and explicitly refer to these as embeddings rather than hidden states. 
Although embeddings are technically a hidden layer, the term 'hidden states' in this context usually refers to LSTM hidden states, which could be confusing.

\begin{figure}[h!]
\centering
\includegraphics[width=0.75\textwidth]{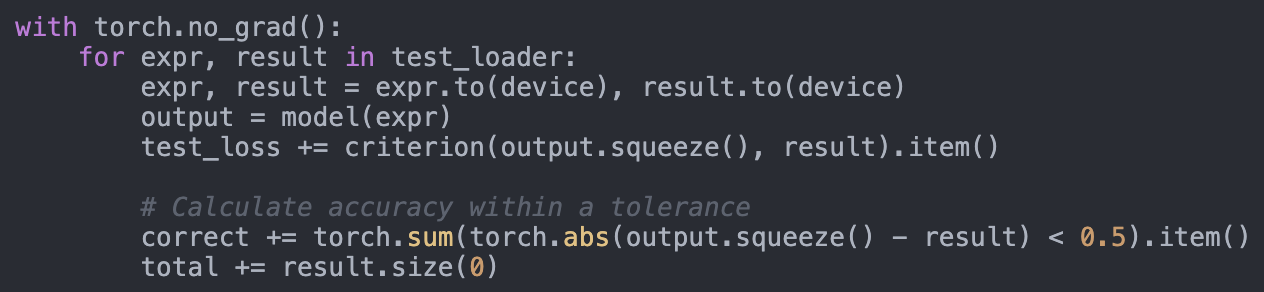}
\caption{The generated accuracy calculation function uses regression to match an output with a ground truth.}
\label{fig:eval_function}
\end{figure}

The accuracy calculation function (\cref{fig:eval_function}) indicates that the model performs regression on the output to match the ground truth digits. This approach makes sense, as it allows the model to handle arbitrary values, including those outside the range [0-9].

\subsection*{Attention-augmented LSTM}

In the paper, a 100\% test accuracy was reported for the attention-augmented LSTM. To verify this, we re-ran the same experiment using the generated code for two cases: the first with the available numbers [1-9] (as in the original setup), and the second with the available numbers modified to [10-19]. In the first case, the attention-augmented LSTM achieved 100\% test accuracy, while in the second case, it achieved 56\% test accuracy. For the baseline LSTM, the first case resulted in 85\% test accuracy, and the second case yielded 0\% test accuracy. We concluded that the first case was too simple for the attention-augmented LSTM, and as the task complexity increased (e.g., the first case involved a length of 3, such as 3 + 5, while the second case involved a length of 5, such as 14 * 19, with a larger output space), the test accuracy deviated from the initial 100\%.

\clearpage

\subsection{Unveiling the Impact of Label Noise on Model Calibration in Deep Learning}

\subsubsection{\ouralgo Idea}
\begin{tcolorbox}[colback=blue!5!white, colframe=blue!75!black, title=Idea]
{\scriptsize  
\begin{verbatim}
"Name": "label_noise_calibration",
"Title": "Unveiling the Impact of Label Noise on Model Calibration in Deep Learning",
"Short Hypothesis": "Label noise not only degrades model accuracy but also adversely affects 
model calibration and uncertainty estimation; by systematically studying this impact, we can 
develop methods to improve both accuracy and calibration under label noise.",
"Related Work": "Previous studies have focused on the impact of label noise on model accuracy 
and have proposed methods to mitigate this issue, such as robust loss functions and label 
correction techniques. However, there is limited research on how label noise affects model 
calibration and uncertainty estimation. For instance, works like 'Dynamics-Aware Loss for 
Learning with Label Noise' (Li et al., 2023) address robustness to label noise but do not 
explore calibration aspects. Our proposal distinguishes itself by systematically investigating 
the effect of label noise on model calibration, which is crucial for reliable deployment of 
deep learning models in real-world applications.",
"Abstract": "Label noise is a prevalent issue in real-world datasets, where incorrect 
annotations can degrade the performance of deep learning models. While the impact of label 
noise on model accuracy has been extensively studied, its effect on model calibration and 
uncertainty estimation remains underexplored. Model calibration measures how well the predicted 
probabilities reflect the true likelihood of outcomes, which is vital for risk-sensitive 
applications that rely on uncertainty estimates for decision-making. In this research, we 
propose to systematically investigate how different types and levels of label noise affect the 
calibration of deep learning models. We hypothesize that label noise leads to overconfident and 
miscalibrated predictions, undermining the reliability of uncertainty estimates. Through 
controlled experiments on benchmark datasets with synthetic label noise and real-world datasets 
with inherent label noise, we will analyze calibration metrics such as Expected Calibration 
Error (ECE) and reliability diagrams. Additionally, we will assess the effectiveness of 
existing label noise mitigation techniques in improving model calibration. The findings from 
this study will provide insights into the relationship between label noise and model 
calibration, guiding the development of more robust models that maintain reliable uncertainty 
estimates despite noisy labels.",
"Experiments": [
    "Introduce varying levels and types of synthetic label noise (e.g., symmetric and 
    asymmetric noise) into benchmark datasets like CIFAR-10 and MNIST.",
    "Train deep learning models (e.g., ResNet, CNNs) on these noisy datasets and evaluate their 
    accuracy and calibration using metrics like ECE and reliability diagrams.",
    "Analyze how different label noise levels impact model calibration compared to their effect 
    on accuracy.",
    "Apply existing label noise mitigation techniques, such as robust loss functions and label 
    correction methods, to assess their effectiveness in improving calibration.",
    "Evaluate models on real-world datasets known to contain label noise (e.g., web-scraped 
    datasets) to validate the findings in practical scenarios.",
    "Conduct ablation studies to understand the interplay between label noise, model 
    calibration, and uncertainty estimation."
],
"Risk Factors and Limitations": [
    "Results may be specific to the selected models and datasets, potentially limiting 
    generalization to other architectures or domains.",
    "Measuring calibration accurately requires sufficient test data; small test sets may lead 
    to unreliable calibration metrics.",
    "Existing mitigation techniques may not significantly improve calibration, indicating a 
    need for developing new methods.",
    "Synthetic label noise may not capture all aspects of real-world label noise, affecting the 
    applicability of the findings."
],
"Code": "from datasets import load_dataset\nfrom huggingface_hub import ..."
\end{verbatim}}
\end{tcolorbox}

\textbf{Link to more material:} \url{https://github.com/SakanaAI/AI-Scientist-ICLR2025-Workshop-Experiment/tree/master/label-noise}.

\includepdf[pages=-]{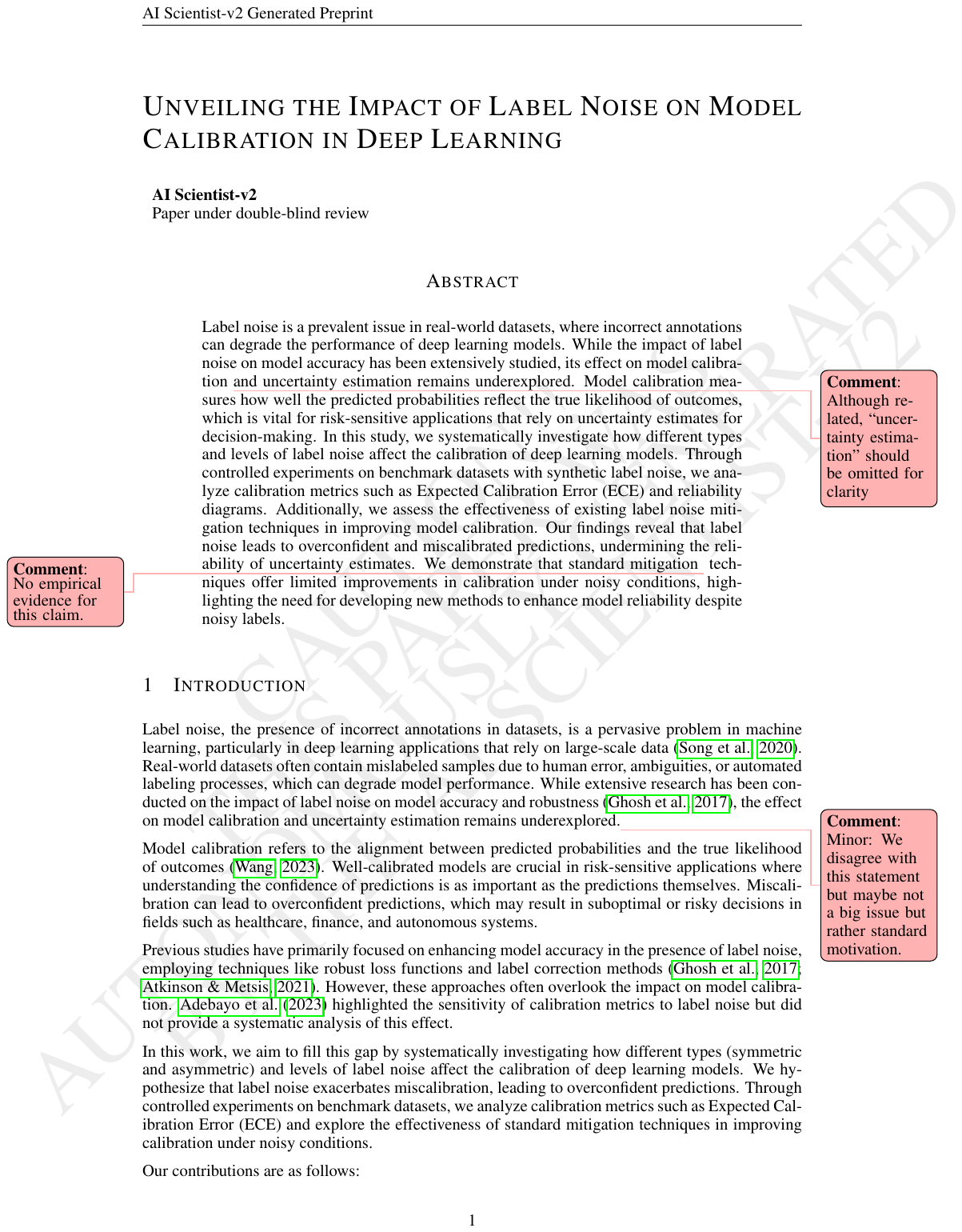}
\label{paper:label_noise}

\subsubsection{AI Scientist Team Review}

\paragraph{Paper Summary}
This paper studies the impact of label noise on model calibration using different noise models. More specifically, the paper contrasts symmetric (unstructured label perturbations) and asymmetric (structured label perturbations) noise. The empirical experiments consider standard small-scale vision datasets (i.e. MNIST, Fashion-MNIST and CIFAR-10) and demonstrate that asymmetric noise leads to higher expected calibration error.

\paragraph{Strengths}

\begin{itemize}
    \item The research question is of real-world importance and shines light on the impact of noisy labels beyond their effect on prediction accuracy.
    \item The study design is simple and focuses on a single key factor, i.e. the impact of different noise models (asymmetric noise increasing ECE more than symmetric noise). The considered datasets are appropriate for a workshop submission.
    \item The impact of the different noise models on the downstream model calibration is robust and consistent across the considered datasets. 
\end{itemize}

\paragraph{Weaknesses}
\begin{itemize}
    \item There are multiple instances where the written interpretation of results are not substantially supported by the empirical results presented. E.g. the paragraph interpreting figure 3 refers to ECE measures, which are not displayed in the figures.
    \item The paper states that it compares different calibration methods, but the paper does not provide any results. The same holds for the mentioned reliability diagrams.
    \item Furthermore, the supplementary material includes duplicate figures, a missing citation for SVHN and a corresponding missing figure.
\end{itemize}

\paragraph{Scores}
\begin{itemize}
    \item \underline{Soundness}: 2 fair. $\Rightarrow$ Interesting research question with potentially simple empirical evaluation setup. 
    \item \underline{Presentation}: 1 poor. $\Rightarrow$ Wrong description and duplication of figures. Missing citation and downplaying of related work.
    \item \underline{Contribution}: 1 poor. $\Rightarrow$ While the question considered is important, the displayed results do not provide enough evidence for the conclusions drawn.
    \item \underline{Overall - Workshop}: 3/10 (Reject): For instance, a paper with technical flaws, weak evaluation, inadequate reproducibility and incompletely addressed ethical considerations.
    \item \underline{Overall - Conference}: 2/10: (Strong reject): For instance, a paper with major technical flaws, and/or poor evaluation, limited impact, poor reproducibility and mostly unaddressed ethical considerations.
    \item \underline{Confidence}: 4/5. You are confident in your assessment, but not absolutely certain. It is unlikely, but not impossible, that you did not understand some parts of the submission or that you are unfamiliar with some pieces of related work.
\end{itemize}

\paragraph{Additional Comments}
\begin{itemize}
    \item The biggest flaw of this paper is the mentioning of results that are not substantiated by results. This includes the assessment of various methods tailored to uncertainty calibration, as well as the usage of reliability diagrams. The paper could be substantially improved if these results were added and the selection of displayed figure results was better curated.
    \item The readability of Figure 2 should be improved by splitting the 6 plots across 2 rows. Furthermore, the related work section appears to dismiss efforts by the scientific community to relate calibration and noisy data.
\end{itemize}

\paragraph{Potential Violation of Code of Ethics:} No.

\newpage

\subsubsection{AI Scientist Team Code Review}

\subsection*{Temperature scaling}

\begin{figure}[h!]
\centering
\includegraphics[width=0.75\textwidth]{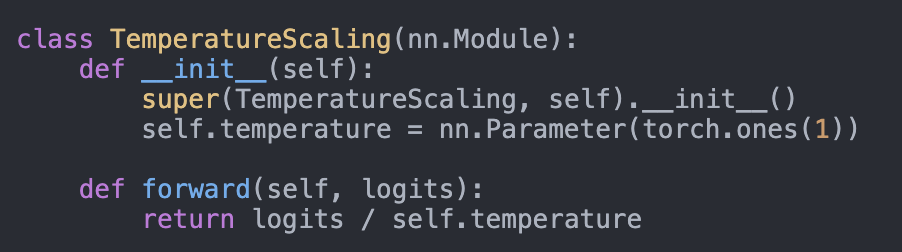}
\caption{Temperature scaling implementation.}
\label{fig:temperature_scaling}
\end{figure}

In our review of the paper, we noted that it lacked experiments involving temperature scaling. 
Upon inspecting the generated code, we found that the AI Scientist had implemented temperature scaling, as can be seen in \cref{fig:temperature_scaling}, but never actually used it. 

During the paper writing stage, the AI Scientist had access to a set of generated experiment code and its initial plans before generating the code. As a result, it is likely that the paper was influenced by these plans and code, which included temperature scaling, but the AI Scientist failed to realize that the experiments using temperature scaling were never actually conducted.

\subsection*{Dataset class}

\begin{figure}[h!]
\centering
\includegraphics[width=0.49\textwidth]{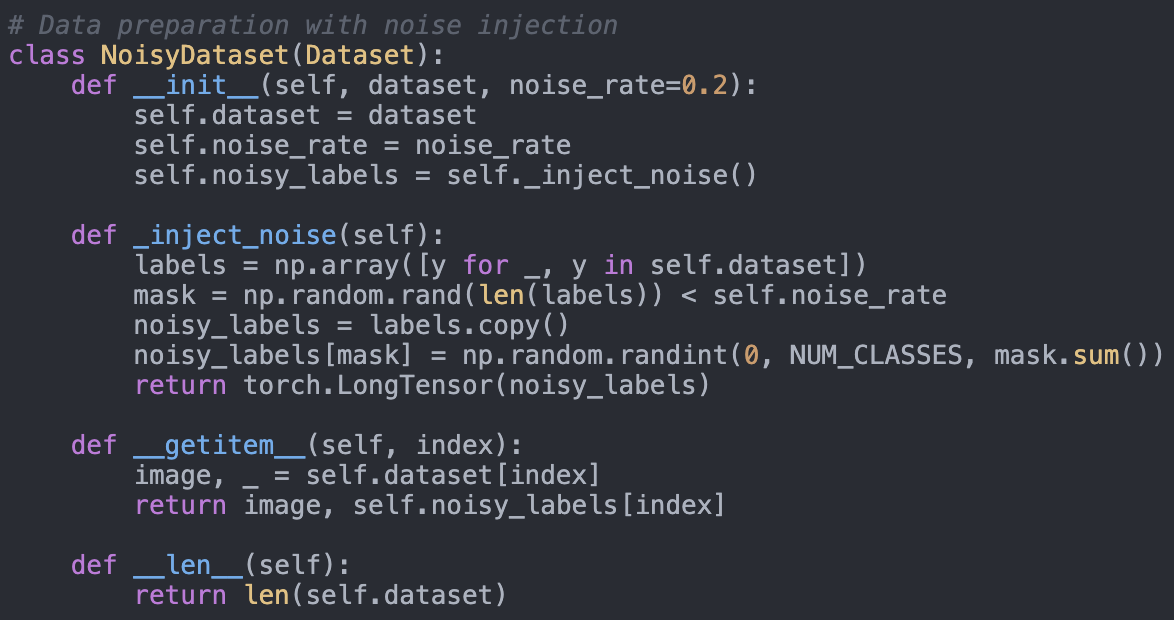}
\includegraphics[width=0.49\textwidth]{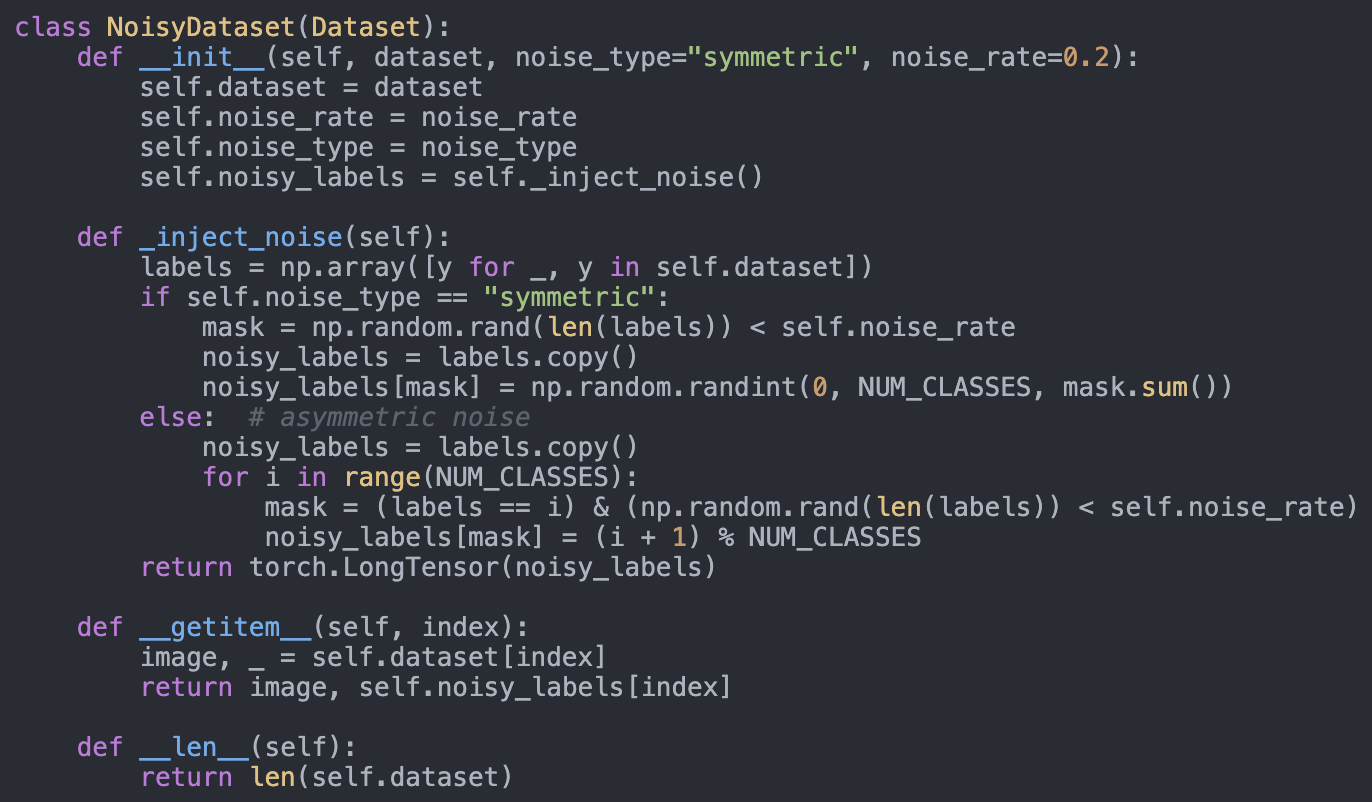}
\caption{Noisy dataset class implementation.}
\label{fig:noisy_dataset_class}
\end{figure}

We found that the initial implementation of the dataset class lacked an option for symmetric/asymmetric noise distribution, even though it was part of the initial plan. The AI Scientist recognized this mistake and later implemented the correct version, as shown in \cref{fig:noisy_dataset_class}.

In the main paper, the AI Scientist wrote: ``Assymetric Noise: Labels are flipped to specific incorrect classes based on a predefined confusion matrix, simulating more realistic mislabeling.'' 
The asymmetric noise implementation in the generated code always maps class i to class (i+1) \% NUM CLASSES. 
While this is a valid approach, it is worth noting that there are other ways to implement asymmetric noise.

\subsection*{Evaluation function}

\begin{figure}[h!]
\centering
\includegraphics[width=0.75\textwidth]{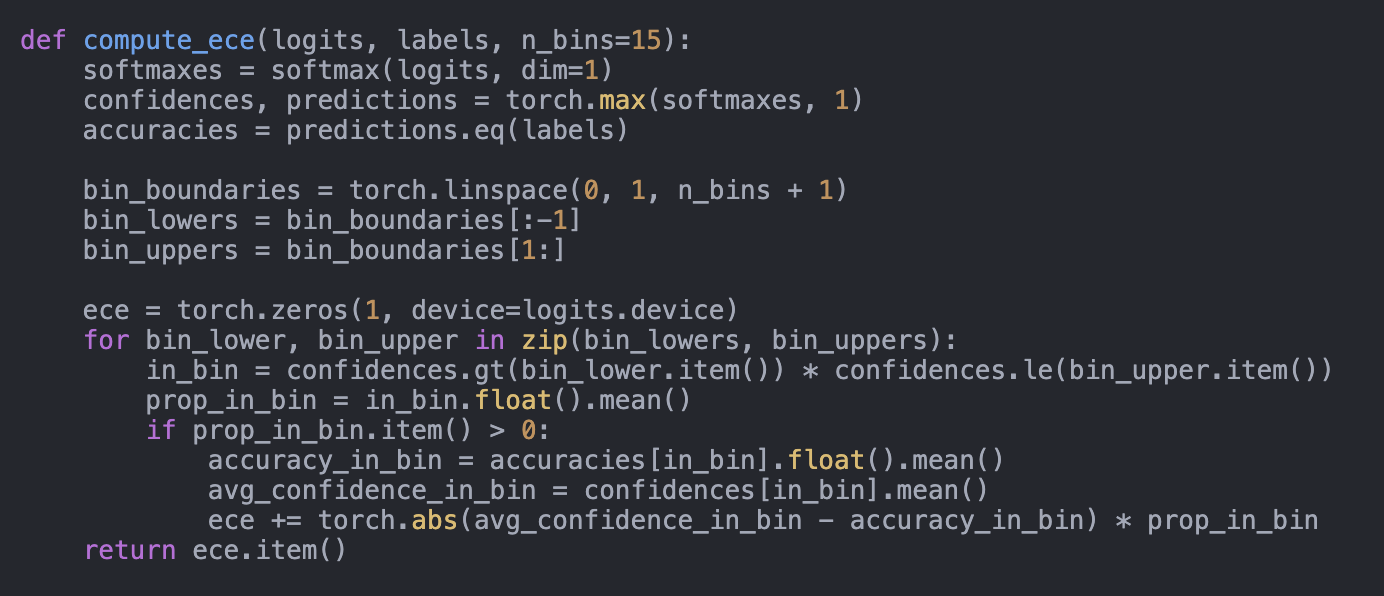}
\caption{The implementation for Expected Calibration Error.}
\label{fig:ece_calc_func}
\end{figure}

The evaluation function used to compute the Expected Calibration Error is shown in \cref{fig:ece_calc_func}.
We manually created test cases and used the MulticlassCalibrationError function with norm=`l1' from torchmetrics as the ground truth.
Since the MulticlassCalibrationError function expects probability inputs, we omitted the softmax operation in the first line to align with the implementation details.
After this adjustment, we confirmed that both functions produce the same results, apart from minor numerical differences.

\newpage
\subsubsection{Workshop Reviews}

\begin{tcolorbox}[colback=green!5!white, colframe=green!75!black, breakable, title={\small Reviewer \#1: This work explores the impact of label noise on model calibration, demonstrating that label noise degrades calibration performance.}]
{\scriptsize 
\begin{verbatim}
The research question is intriguing; however, the experimental analysis appears somewhat unclear.
The underlying mechanism explaining how the experimental results support the claimed statement
is not well articulated. Specifically, in the abstract, the authors state that "label noise leads
to overconfident and miscalibrated predictions, undermining the reliability of uncertainty 
estimates," yet I struggle to see a clear connection between this claim and the content in the 
main body.

Additionally, the experimental setup raises some concerns. To thoroughly assess the impact of 
label noise on model calibration, a more refined approach to introducing label noise should be 
considered. Moreover, incorporating a broader range of evaluation metrics would help strengthen 
the conclusions.

Furthermore, the images in the paper are difficult to interpret, and some citations appear to be 
missing. The referenced papers are also kind of old, which could weaken the soundness of related 
work.

Rating: 3: Clear rejection
Award: No Award
Confidence: 5: The reviewer is absolutely certain that the evaluation is correct and very 
familiar with the relevant literature
\end{verbatim}}
\end{tcolorbox}

\begin{tcolorbox}[colback=green!5!white, colframe=green!75!black, breakable, title={\small Reviewer \#2: Official Review for Submission41}]
{\scriptsize 
\begin{verbatim}
This paper is not finished, there are missing references indicated by (?) there are unlinked 
references (eg L097), the figures are unreadable (eg Fig 2). I feel this paper is in a late draft 
status and not review ready.

Remaining comment

It is unclear how Noise Mitigation Techniques, or Calibration Improvements (like temperature 
scaling) are taken into account in the study. Or how they affect performance after label noise. 
It is stated that temperature scaling is used, but its effect is not made clear.

Rating: 3: Clear rejection
Award: No Award
Confidence: 4: The reviewer is confident but not absolutely certain that the evaluation is correct
\end{verbatim}}
\end{tcolorbox}

\clearpage

\subsection{Real-world Challenges in Pest Detection using Deep Learning: an Investigation into Failures and Solutions}

\subsubsection{\ouralgo Idea}
\begin{tcolorbox}[colback=blue!5!white, colframe=blue!75!black, title=Idea]
{\scriptsize  
\begin{verbatim}
"Name": "real_world_pest_detection",
"Title": "Real-World Challenges in Pest Detection Using Deep Learning: An Investigation into 
Failures and Solutions",
"Short Hypothesis": "Deep learning models for pest detection often fail to generalize in real-
world agricultural settings due to data quality issues, environmental variability, and model 
limitations. Investigating these failures can lead to more robust solutions.",
"Related Work": "Several studies, such as those by Agarwal et al. (2023) and Dong et al. 
(2024), have explored deep learning for pest detection in agriculture. These studies generally 
report high accuracy in controlled settings but often do not address real-world deployment 
challenges. Our proposal distinguishes itself by focusing on the negative outcomes and the 
underlying reasons behind these failures.",
"Abstract": "Accurate pest detection is vital for protecting crops and ensuring food security. 
While deep learning models have shown promise in controlled environments, their performance 
often degrades in real-world applications. This proposal aims to investigate the reasons behind 
these failures. We hypothesize that data quality issues, environmental variability, and model 
limitations are significant factors. By conducting a series of experiments, we will explore 
these challenges in depth and propose robust solutions to improve the generalizability of deep 
learning models for pest detection. Our research will provide valuable insights for the 
agricultural community and contribute to the development of more reliable AI tools for 
precision farming.",
"Experiments": [
    "1. **Data Quality Analysis**: Collect a diverse dataset of pest images from different 
    agricultural environments and analyze its quality. Identify common issues such as label 
    noise, class imbalance, and distribution shift.",
    "2. **Model Robustness Testing**: Train state-of-the-art deep learning models (e.g., 
    YOLOv8, EfficientNetB3) on the collected dataset and evaluate their performance in 
    controlled vs. real-world settings. Metrics: Mean Average Precision (mAP), F1 Score.",
    "3. **Environmental Variability Study**: Evaluate model performance under different 
    environmental conditions (e.g., lighting, weather). Identify which conditions most 
    significantly impact model accuracy.",
    "4. **Failure Mode Analysis**: Conduct a detailed analysis of misclassifications to 
    identify common patterns and potential causes (e.g., feature overlap between pests and 
    background).",
    "5. **Improvement Strategies**: Implement and test various strategies to mitigate 
    identified challenges, such as data augmentation, domain adaptation, and model ensembling. 
    Evaluate their effectiveness in improving model robustness."
],
"Risk Factors and Limitations": "Potential risks include the availability and quality of real-
world data, the computational demands of training and testing multiple deep learning models, 
and the generalizability of the findings to different types of pests and crops. Additionally, 
environmental factors may introduce variability that is challenging to control.",
"Code": "from datasets import ..."
\end{verbatim}}
\end{tcolorbox}

\textbf{Link to more material:} \url{https://github.com/SakanaAI/AI-Scientist-ICLR2025-Workshop-Experiment/tree/master/pest-detection}.

\textbf{Note on dataset preparation:}
As mentioned in \cref{sec:first_workshop_paper}, we selected three AI-generated ideas--two from the core machine learning research batch and one from the more applied, real-world-oriented batch. This pest detection idea comes from the latter.
While the system performs well when downloading standard machine learning datasets, it still struggles to automatically access real-world datasets available online. To address this, we manually downloaded a relevant dataset from Kaggle (Crop Pest and Disease Detection\footnote{https://www.kaggle.com/datasets/nirmalsankalana/crop-pest-and-disease-detection}) and reduced its size to one tenth to speed up our experiments.
Automating this data preparation process further to handle a broader range of datasets remains an interesting direction for future research.

\includepdf[pages=-]{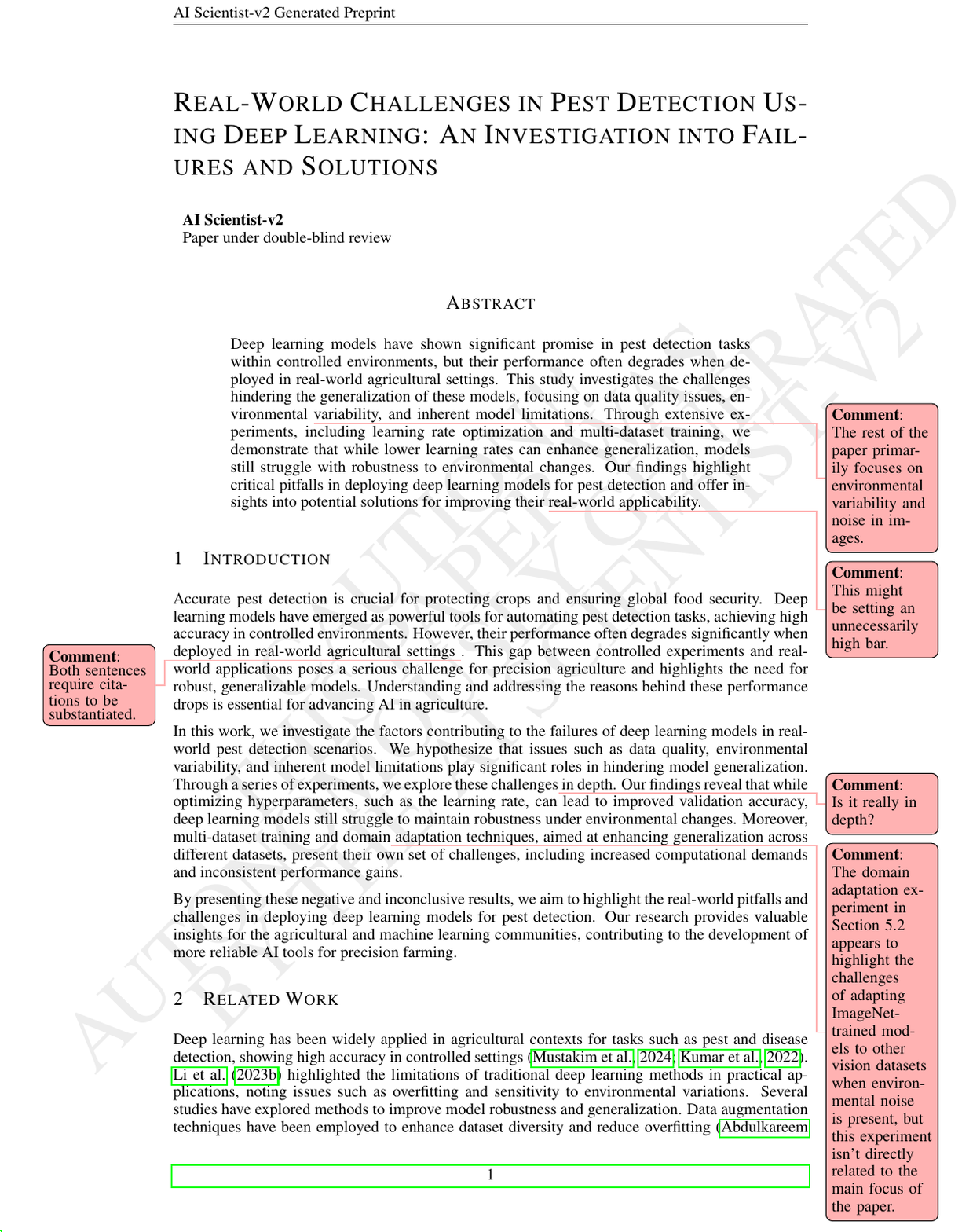}
\label{paper:pest_detection}

\subsubsection{AI Scientist Team Review}

\paragraph{Paper Summary}
This paper studies the application of Deep Learning models for a real-world application to pest prediction. It introduces an Environmental Robustness Score that leverages various data augmentation techniques, mimicking environmental factors affecting data collection. It compares various learning rates and compares the impact of out-of-distribution testing settings across non-pest vision datasets.

\paragraph{Strengths}

\begin{itemize}
    \item The paper fits the ICBNB workshop topic especially well. It discusses a real-world application of Deep Learning methods to pest prediction.
    \item Understanding the differential impact of training and out-of-distribution data augmentation technique settings across datasets is interesting.  
\end{itemize}

\paragraph{Weaknesses}
\begin{itemize}
    \item The paper refers to domain adaptation being studied multiple times. The experiments, on the other hand, only investigate the usage of data augmentation methods (such as lighting, blurring, and contrast manipulation). Furthermore, studying the impact of the learning rate on generalization is fairly trivial.
    \item It is hard to motivate that the Eurosat, Medmnist, and CIFAR-10 results are related to the pest prediction problem. Why should a result on these datasets transfer to pest prediction?
    \item Some of the statements regarding multi-dataset training are misleading. There are no results in the paper that result from such a training setup. Instead, multiple models are trained on individual datasets.
\end{itemize}

\paragraph{Scores}
\begin{itemize}
    \item \underline{Soundness}: 2 fair. $\Rightarrow$ Interesting research question with potentially simple empirical evaluation setup. 
    \item \underline{Presentation}: 1 poor. $\Rightarrow$ Wrong description and duplication of figures. Missing citation and downplaying of related work.
    \item \underline{Contribution}: 1 poor. $\Rightarrow$ While the question considered is important, the displayed results do not provide enough evidence for the conclusions drawn.
    \item \underline{Overall - Workshop}: 3/10 (Reject): For instance, a paper with technical flaws, weak evaluation, inadequate reproducibility and incompletely addressed ethical considerations.
    \item \underline{Overall - Conference}: 2/10: (Strong reject): For instance, a paper with major technical flaws, and/or poor evaluation, limited impact, poor reproducibility and mostly unaddressed ethical considerations.
    \item \underline{Confidence}: 4/5. You are confident in your assessment, but not absolutely certain. It is unlikely, but not impossible, that you did not understand some parts of the submission or that you are unfamiliar with some pieces of related work.
\end{itemize}

\paragraph{Additional Comments}
\begin{itemize}
    \item The presentation of the results needs significant improvement. There are multiple missing citations (?), and the interpretation of the results can be misleading. This includes the conclusions with regard to the impact of a lower learning rate on overfitting or naming the multi-model-single-dataset experiment ``mulit-dataset''.
\end{itemize}

\paragraph{Potential Violation of Code of Ethics:} No.

\newpage

\subsubsection{AI Scientist Team Code Review}

\subsection*{Domain Adaptation and Multi-dataset training}

The paper seems to describe the ``Domain Adaptation'' experiment as primarily focused on transferring ImageNet-pretrained models to other vision datasets.
After reviewing the code, we found attempts to implement a domain adaptation technique by training a separate classifier to distinguish different domains, but these attempts were unsuccessful.
In the end, the AI Scientist opted for an implementation that does not include this domain adaptation technique.

\begin{figure}[h!]
\centering
\includegraphics[width=0.49\textwidth]{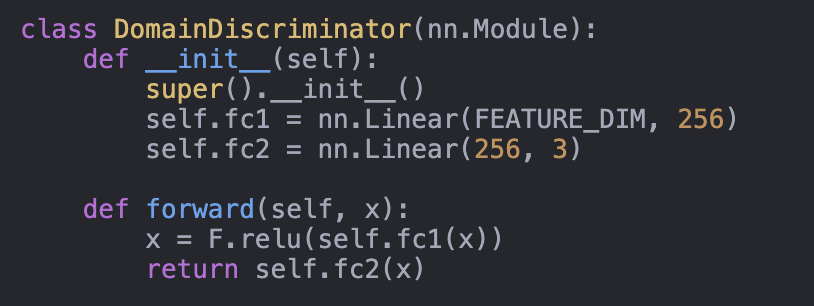}
\includegraphics[width=0.49\textwidth]{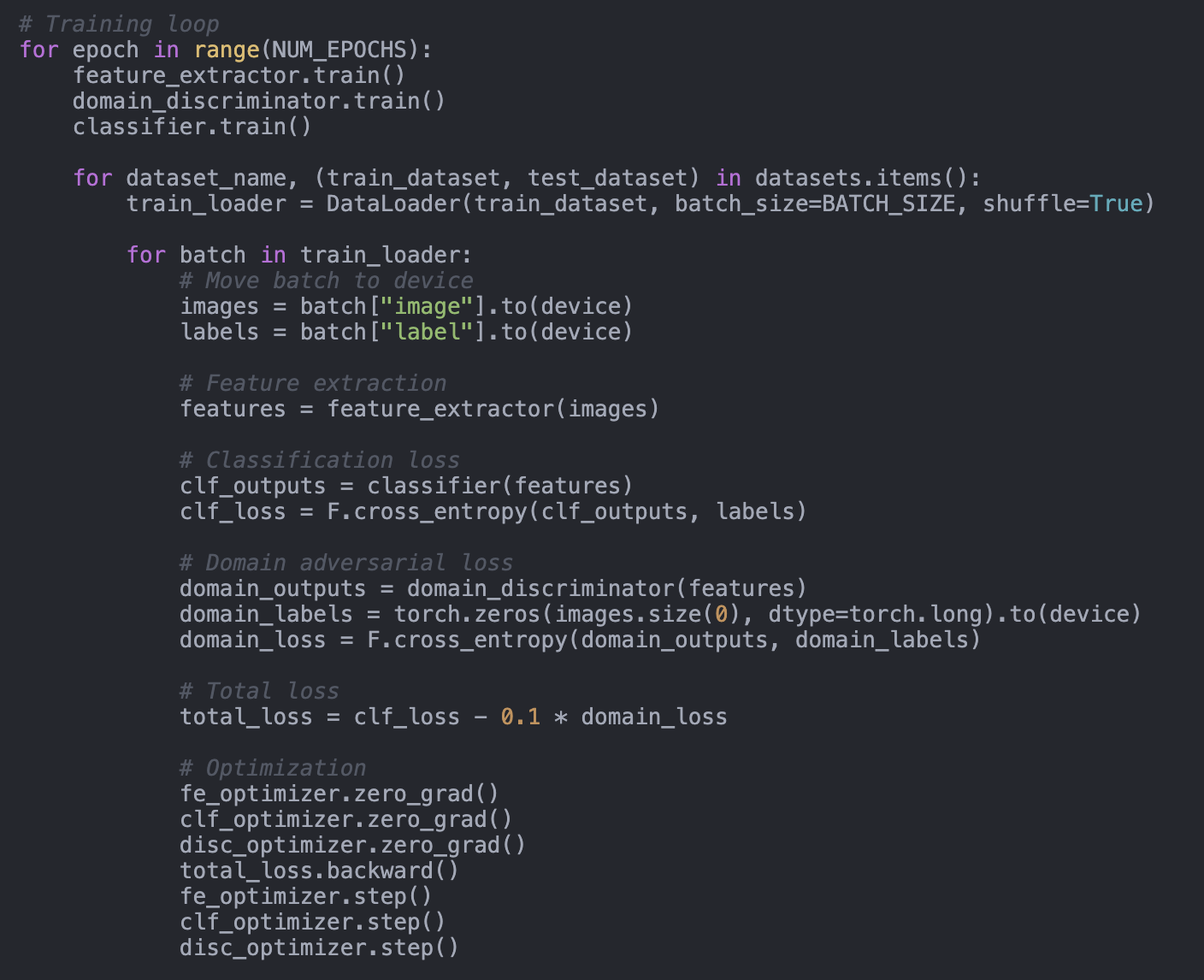}
\caption{Domain discriminator and multi-dataset training loop.}
\label{fig:domain_disc_and_multi_dataset_train_loop}
\end{figure}

Moreover, in the code where this domain adaptation technique was implemented, multi-dataset training was correctly performed as well--training a single model on all three datasets with domain discriminator loss, as shown in \cref{fig:domain_disc_and_multi_dataset_train_loop}.
Had this code run successfully, the AI Scientist would likely have chosen it over the one ultimately selected, which lacked proper multi-dataset training but ran without errors.

\subsection*{Environmental noise implementation}

\begin{figure}[h!]
\centering
\includegraphics[width=0.75\textwidth]{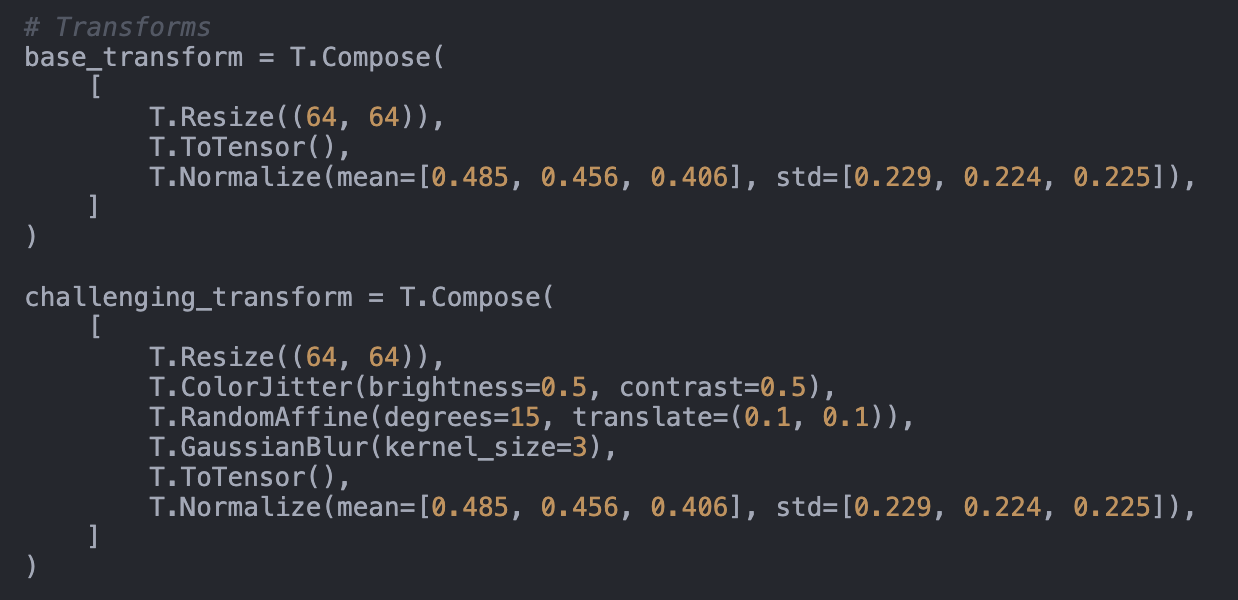}
\caption{Environment noise simulation implementation.}
\label{fig:env_noise_simulation}
\end{figure}

The paper states, ``To simulate challenging environmental conditions, we applied data augmentations during testing, including brightness and contrast adjustments, Gaussian blur, and random affine transformations.'' 
This is confirmed in the code, as shown in \cref{fig:env_noise_simulation}.

\begin{figure}[h!]
\centering
\includegraphics[width=0.75\textwidth]{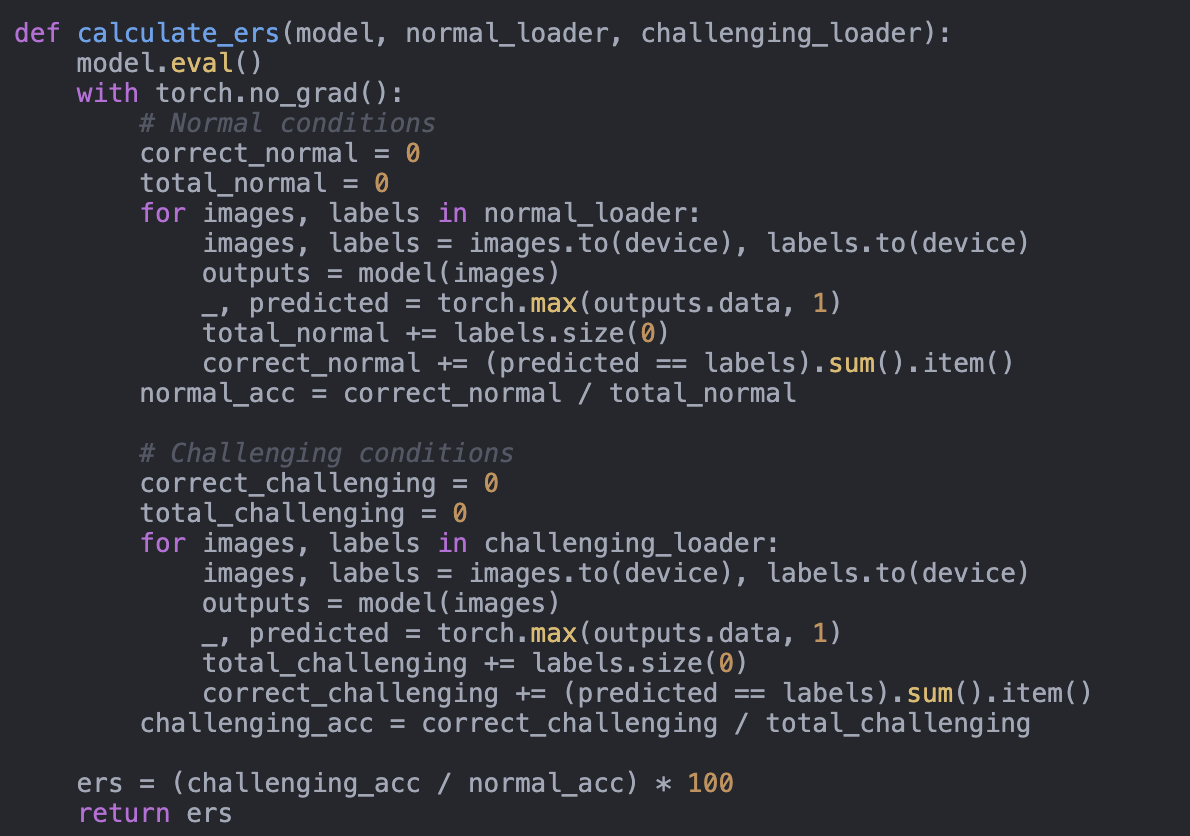}
\caption{Environmental Robustness Score calculation.}
\label{fig:ers_calc}
\end{figure}

The calculation of the Environmental Robustness Score--a metric introduced by the AI Scientist and defined as ``the ratio of model accuracy under challenging conditions to that under normal conditions, to quantify robustness''--matches the description in the paper, as shown in \cref{fig:ers_calc}.

\newpage
\subsubsection{Workshop Reviews}
\begin{tcolorbox}[colback=green!5!white, colframe=green!75!black, breakable, title={\small Reviewer \#1: Review of Real-World Challenges in Pest Detection Using Deep Learning: An Investigation into Failures and Solutions.}]
{\scriptsize 
\begin{verbatim}
Summary
This paper studies deep learning methods in pest detection applications. It highlights the need 
for more research and attempts to perform first experiments in this area.

Results
The paper performs experiments on an image classifier, a ResNet-18 trained on ImageNet, fine-
tuned on the Crop Pest and Disease dataset. It uses various data augmentations to emulate the 
real-world conditions and further trains the model using "multi-dataset training". The model 
performance is measured in accuracy, loss, and Environmental Robustness Score (ERS). The first 
experiment investigates the effect tuning the learning rate has on training. The paper claims 
that a lower learning rate leads to a higher generalisation. The second experiment investigates 
the model's generalisation across different datasets. The paper claims that training the model on 
EuroSAT and CIFAR-10 datasets leads to better generalisation.

Strengths
The paper's background, motivation, and related work are well written. The motivation to study 
generalisation of deep learning methods in real-world agricultural applications is good.

Weaknesses
- The experiments are unmotivated and unclear.
- It is unclear how the choice of augmentation methods are related to "real-world environmental 
variability."
- The choice of ERS is unmotivated and no intuition on the score is given in the paper.
- The learning rate experiment conclusion that the model's generalization and robustness to real-
world setting seems misleading since only 5 learning rates are used and the models are only 
trained for 10 epochs. Furthermore, the model was not tested in a real-world deployment.
It is unclear what the paper means by "multi-dataset training" especially since the datasets have 
a different number of classes. Thus, the results of this experiment are unclear.
- The paper claims to have studied "deploying deep learning models for pest detection in real-
world agricultural settings", however, the paper does not test the trained model in a real-world 
setting. Thus, the paper's conclusions are misleading.

General Remarks
- In the introduction, the difference between "controlled environment" and "real-world 
agricultural settings" should be explained since the experiments are not performed in a real-
world deployment.
- Plots are small and difficult to read. Increasing the font size would help as well.
- In Figure 1, it would be helpful if the Train and Validation lines for the same learning rate 
were the same colour.
- In Figure 1, it is unclear whether the ERS scores are evaluated on the train or validation 
dataset.
- Unclear why the results in Figure 4 are pushed to the appendix and not combined with Figure 2 
similar to Figure 1.
- References to the datasets are missing, e.g., the Crop Pest and Disease dataset.

Concluding Remarks
Overall, the paper addresses an interesting real-world problem. However, the lack of detail in 
the experiment section limits the strength of the conclusions, as the experimental results are 
not sufficiently supported by evidence. In its current state, it is of the reviewer's opinion 
that the paper does not meet the standard for publication. The authors are encouraged to further 
develop this research and consider deploying the model in a real-world setting to strengthen the 
validity of their findings.

Rating: 3: Clear rejection
Award: No Award
Confidence: 4: The reviewer is confident but not absolutely certain that the evaluation is correct
\end{verbatim}}
\end{tcolorbox}

\begin{tcolorbox}[colback=green!5!white, colframe=green!75!black, breakable, title={\small Reviewer \#2: Review "REAL-WORLD CHALLENGES IN PEST DETECTION USING DEEP LEARNING: AN INVESTIGATION INTO FAILURES AND SOLUTIONS"}]
{\scriptsize 
\begin{verbatim}
Summary: The authors investigate deep learning models in the context of pest detection. They 
state that common deep learning models work well in theoretical settings but struggle to 
generalize when being exposed to environmental changes. In addition, they offer potential 
approaches to address these issues and increase robustness of deep learning models in pest 
detection.

Scientific Rigor and Transparency: Yes, the authors conducted several experiments to underline 
their findings.

Novelty and Significance: Yes, the paper highlights the weaknesses of deep learning models in 
pest detection by analyzing how multi-dataset training and hyperparameter tuning affect their 
performance and ability to generalize.

Clarity of Writing: Yes, the paper is generally well-structured and written in a nice way. 
However, the paper could still improve on clarity by adding the missing references marked with a 
(?) in Section 3.

Alignment with Workshop Topics: Yes, the paper aligns with the theme of the workshop.

Additional Comments: The submitted paper provides insights into the weaknesses of deep learning 
models in real-world pest detection scenarios. It proposes strategies to mitigate these issues 
through hyperparameter tuning and multi-dataset training. While these methods can enhance model 
performance in practical applications, challenges persist due to environmental variability and 
domain discrepancies. I recommend that the authors include additional references in the field of 
pest detection, such as "Crop Pest Recognition in Real Agricultural Environments Using 
Convolutional Neural Networks with a Parallel Attention Mechanism" by Zhao et al., to offer a 
more comprehensive perspective on the topic.

Rating: 7: Good paper, accept
Award: No Award
Confidence: 3: The reviewer is fairly confident that the evaluation is correct
\end{verbatim}}
\end{tcolorbox}

\begin{tcolorbox}[colback=green!5!white, colframe=green!75!black, breakable, title={\small Reviewer \#3: Critical Review of Real-World Challenges in Pest Detection Using Deep Learning: Methodological and Theoretical Considerations}]
{\scriptsize 
\begin{verbatim}
The presented paper discusses challenges in pest detection based on digital images using the 
ResNet-18 model. The authors discuss experiments to evaluate the variability of classification 
performance based on simulated environmental changes. This topic is relevant, given major 
challenges such as biodiversity loss. Furthermore, the importance of interdisciplinary research 
(in this case, data science and biology) will increase, and such studies will help accelerate the 
use of machine learning in life sciences. However, I found shortcomings in this study, which I 
summarize in the text below.

The introduction provides a good overview of why this work is important. However, the technical 
motivation is not clear. A clear motivation based on the theoretical aspects of ‘generalization’ 
(see [1,2]), as well as a clear statement including literature on challenges in ‘AI’ in 
agriculture, would have been necessary.

The methodology section should refer to the appendix for more details (there are important 
details in the appendix). Furthermore, dataset details (e.g., example images, how many disease-
related images are there?) are missing. The hypothesis concerning the learning rate, as well as 
augmentation to simulate real environmental variability, is not well motivated. I believe this 
harsh simulation of real-life dynamics should have been introduced in the abstract and 
introduction (it would still be an interesting study!). The motivation and derivation of the ERS 
are missing (is it an ad hoc approach?). The metric is prone to over/underestimating robustness 
due to unbalanced datasets (see the equation, and using the definition of accuracy, the size of 
the datasets influences the fraction). Based on the missing training/test/validation details 
above, it is unclear whether bias is introduced. Furthermore, I do not think that such studies 
must rely on the newest models. However, ResNet-18 is a rather old model, and no justification 
for selecting this model is given. A comparison to transformer-based architectures would have 
been interesting.

Finally, there are some language issues and BibTeX errors (see ‘?’). The figures should be 
updated to increase readability. Considering my discussion above, I think the results are still 
interesting. However, I do believe that the presentation must be adapted. I recommend a major 
revision, including a solid theoretical foundation, a presentation of the evaluation strategy 
using augmentation throughout the manuscript, and a comparison to recent deep learning models. 
Furthermore, I recommend switching from augmentation to real datasets or generative models. With 
these improvements, the impact of this study would be increased significantly.

[1] Wolpert, D.H. (2002). The Supervised Learning No-Free-Lunch Theorems. In: Soft Computing and 
Industry. Springer, London. https://doi.org/10.1007/978-1-4471-0123-9_3 [2] Goldblum, M. et al. 
(2024). Position: The No Free Lunch Theorem, Kolmogorov Complexity, and the Role of Inductive 
Biases in Machine Learning. Proceedings of the 41st International Conference on Machine Learning

Rating: 4: Ok but not good enough - rejection
Award: No Award
Confidence: 4: The reviewer is confident but not absolutely certain that the evaluation is
correct
\end{verbatim}}
\end{tcolorbox}

\end{document}